\documentclass[conference]{IEEEtran}
\usepackage{times}

\usepackage[numbers, sort&compress]{natbib}
\usepackage{multicol}
\usepackage[bookmarks=true]{hyperref}
\usepackage{amsmath}
\usepackage{graphicx}
\usepackage{booktabs}
\usepackage{multirow}
\usepackage{cuted}
\usepackage{capt-of}
\usepackage{placeins}
\usepackage{colortbl}
\usepackage[table]{xcolor}

\usepackage{amssymb}
\usepackage{algorithm}
\usepackage{algpseudocode}
\usepackage{float}

\usepackage{hyperref}
\usepackage{url}
\usepackage{xcolor}
\hypersetup{
    colorlinks=true,
    linkcolor=black,
    filecolor=magenta,      
    urlcolor=magenta,
}

\usepackage{caption}
\begin{document}

\title{Rhythm: Learning Interactive Whole-Body Control for Dual Humanoids}

\author{
    \IEEEauthorblockN{
        Hongjin Chen\textsuperscript{1,2,*} \quad
        Wei Zhang\textsuperscript{2,*} \quad
        Pengfei Li\textsuperscript{2,3,\textdagger} \quad
        Shihao Ma\textsuperscript{1,2} \quad
        Ke Ma\textsuperscript{1,2} \quad
        Yujie Jin\textsuperscript{2} \quad
        Zijun Xu\textsuperscript{1,5} \\
        Xiaohui Wang\textsuperscript{1,2} \quad
        Yupeng Zheng\textsuperscript{6} \quad
        Zining Wang\textsuperscript{2} \quad
        Jieru Zhao\textsuperscript{4} \quad
        Yilun Chen\textsuperscript{2} \quad
        Wenchao Ding\textsuperscript{1,2,\textdagger}
    }
    \vspace{0.25em}
    \IEEEauthorblockA{
        \textsuperscript{1}Fudan University \quad
        \textsuperscript{2}TARS Robotics \quad
        \textsuperscript{3}Tsinghua University \quad
        \textsuperscript{4}Shanghai Jiao Tong University \\
        \textsuperscript{5}Shanghai Innovation Institute \quad
        \textsuperscript{6}Institute of Automation, Chinese Academy of Sciences
    }
    \vspace{0.25em}
    \IEEEauthorblockA{
        \textsuperscript{*}Equal contribution \qquad
        \textsuperscript{\textdagger}Corresponding Authors
    }
    \vspace{0.25em}
    \IEEEauthorblockA{
        Project Page: \url{https://hoshi-no-ai.github.io/Rhythm/}
    }
}

\maketitle

\begin{strip}
    \vspace{-1.5cm} 
    \centering
    \includegraphics[width=\textwidth]{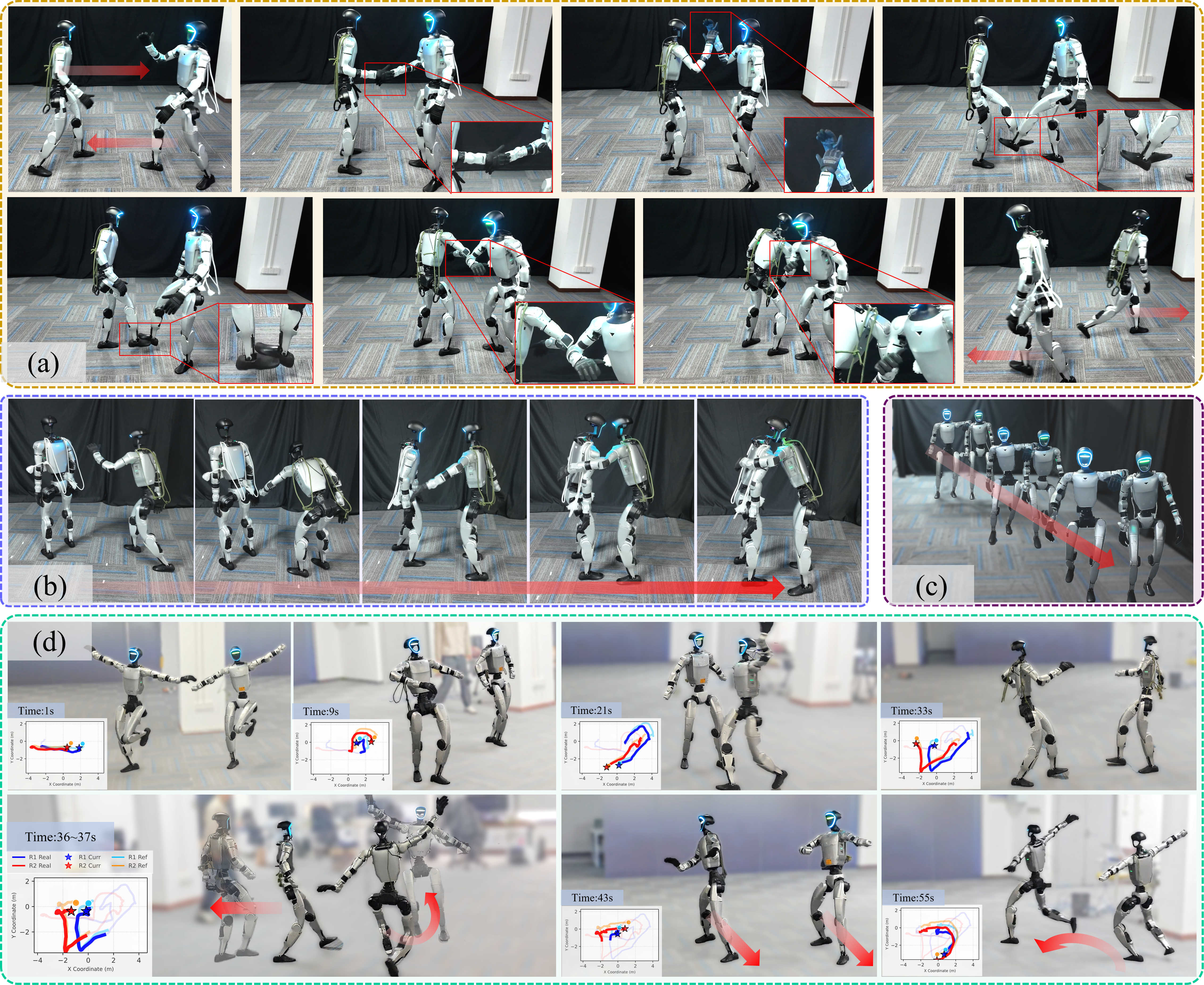}

    \vspace{-0.1cm} 

    \captionof{figure}{The proposed framework, \textbf{Rhythm}, facilitates a spectrum of humanoid–humanoid interactions.
    \textbf{(a–c) Contact-Rich Interaction:} The method handles interactions ranging from light contact (Greeting) to intensive contact (Hug, Shoulder-to-Shoulder), maintaining fine-grained contact geometry without penetration (shown in the zoomed-in views).
    \textbf{(d) Coordinated Interaction:} The humanoids perform synchronized long-horizon dance (\textit{La La Land}), with trajectories showing consistent spatiotemporal alignment and stable relative positioning over time.}
    
    \label{fig:teaser}

    \vspace{-0.3cm}
\end{strip}

\begin{abstract}

Realizing interactive whole-body control for multi-humanoid systems is critical for unlocking complex collaborative capabilities in shared environments. Although recent advancements have significantly enhanced the agility of individual robots, bridging the gap to physically coupled multi-humanoid interaction remains challenging, primarily due to severe kinematic mismatches and complex contact dynamics. To address this, we introduce Rhythm, the first unified framework enabling real-world deployment of dual-humanoid systems for complex, physically plausible interactions. Our framework integrates three core components: (1) an Interaction-Aware Motion Retargeting (IAMR) module that generates feasible humanoid interaction references from human data; (2) an Interaction-Guided Reinforcement Learning (IGRL) policy that masters coupled dynamics via graph-based rewards; and (3) a real-world deployment system that enables robust transfer of dual-humanoid interaction. Extensive experiments on physical Unitree G1 robots demonstrate that our framework achieves robust interactive whole-body control, successfully transferring diverse behaviors such as hugging and dancing from simulation to reality.

\end{abstract}

\IEEEpeerreviewmaketitle

\section{Introduction}

Humanoid robotics has witnessed rapid evolution, achieving remarkable milestones in single-agent capabilities. Recent research has established a strong foundation in dynamic locomotion \cite{wang2025beamdojo, li2025gaitnet, xue2025unified, gu2024advancing, huang2025standing} and general whole-body control \cite{li2025amo, cheng2024expressive, liao2025beyondmimic, he2025asap, zhang2025track, shao2025langwbc}, significantly enhancing the agility and robustness of individual robots. However, the broader vision of embodied intelligence necessitates agents that can operate beyond isolation \cite{sheng2025comprehensive, cao2025humanoid}. Realizing multi-agent systems capable of physical collaboration represents a critical next step. Yet, research in this domain remains disproportionately focused on single-robot tasks.

Despite growing interest in multi-agent interaction, current solutions are largely confined to virtual environments or simplified humanoid-object interactions. 
In computer graphics, physics-based animation has achieved realistic simulations of multi-character interactions \cite{liang2024intergen,xu2024interx,yao2025physiinter,gao2024coohoi,zhang2023simulation}, yet these methods often prioritize visual fidelity over the strict physical constraints necessary for real-world robotic deployment.

In robotics, although human–robot collaboration \cite{du2025learning, huang2026learning} and competitive sports in structured environments \cite{su2025hitter, liu2025humanoid, xu2025learning} have been explored, these typically involve a compliant human partner or passive objects.
Research explicitly targeting multi-humanoid interaction remains scarce, and existing works are predominantly validated only in simulation \cite{liu2025takes}. Consequently, achieving robust, physically coupled whole-body control on real multi-humanoid hardware remains an unbridged gap in the field.

Two fundamental challenges hinder the Sim-to-Real transition for dual-humanoid systems: (1) the scarcity of feasible interaction data; and (2) the complexity of the training-to-deployment paradigm. 
First, acquiring high-quality interaction references is non-trivial. Directly transferring \textit{Human-Human Interaction} \cite{wang2026intermoe, xu2025perceiving, starke2020local,xu2024interx} data to robots introduces severe kinematic conflicts due to morphological differences between humans and humanoids (see Sec.~\ref{subsec:retargeting}). Standard retargeting methods \cite{araujo2025retargeting, luo2023perpetual, yang2025omniretarget} struggle to preserve both individual motion style and precise interaction geometry, yielding suboptimal motion references. Second, the learning and deployment pipeline presents significant hurdles. Existing tracking policies \cite{liao2025beyondmimic, ze2025twist, xie2025kungfubot} typically treat agents as isolated entities, failing to model the intricate coupled dynamics essential for close interaction.
Meanwhile, a significant disparity exists between the global observability available in simulation and the asynchronous, ego-centric, partially observable reality of real hardware, making the deployment unstable.

To address these challenges, we introduce \textbf{Rhythm} (Inte\underline{r}active W\underline{h}ole-Bod\underline{y} Con\underline{t}rol for Dual \underline{H}u\underline{m}anoids), a unified framework designed to empower dual humanoids to execute complex, physically plausible interactions in real-world scenarios. The framework explicitly models interaction geometry and physical contact to achieve high-fidelity coupled behaviors. 
First, to resolve kinematic conflicts in reference generation, we introduce \textbf{Interaction-Aware Motion Retargeting (IAMR)}. By explicitly modeling interaction geometry and utilizing distance-aware dynamic weighting, this module adaptively balances self-motion fidelity with interaction geometry. The resulting geometrically consistent references serve as a crucial prior, laying the physical foundation for the subsequent training phase. Building upon this, we develop \textbf{Interaction-Guided Reinforcement Learning (IGRL)} to master the complex dual-agent dynamics. This module directly leverages the interaction geometry and contact preserved by IAMR through explicit graph-based rewards, guiding agents to learn synchronized and robust behaviors. Finally, to realize dual-humanoid interactive whole-body control in the real world, we implement a relative state estimation and inter-agent synchronization scheme, effectively bridging the gap between global simulation and ego-centric reality.

Fig.~\ref{fig:teaser} illustrates the robustness of our framework in real-world scenarios, successfully handling tasks ranging from intensive contacts to synchronized long-horizon dancing. Our main contributions are summarized as follows:
\begin{itemize}
    \item We propose \textbf{Rhythm}, a unified framework for whole-body dual-humanoid interaction that, to our knowledge, achieves the first successful \textbf{robust transfer} of complex interactive behaviors to physical hardware.
    
    \item We develop \textbf{IAMR} to resolve kinematic conflicts and generate humanoid-humanoid interaction motion references. Furthermore, we release \textbf{MAGIC}, the \underline{M}ulti-Hum\underline{a}noid \underline{G}eometric \underline{I}ntera\underline{c}tion Dataset, offering paired raw and retargeted interaction data.
    
    \item We introduce \textbf{IGRL}, a multi-agent learning module that masters coupled interaction dynamics. By incorporating graph-based rewards, it enables agents to learn robust, physically consistent interactive behaviors.
    
    \item We conduct extensive experiments on Unitree G1 humanoids in simulation and on physical hardware. Quantitative and qualitative evaluations across diverse interaction tasks demonstrate the superior performance and robustness of our framework.
\end{itemize}

\section{Related Work}

\subsection{Learning-Based Humanoid Control and Interaction} Learning-based whole-body control methods primarily leverage Reinforcement Learning (RL) to endow humanoid robots with diverse motor skills \cite{he2025gettingup, huang2025standing, he2025asap, xue2025unified, wang2025beamdojo, ben2025homie, allshire2025visual, zhang2025hub, peri2025nonconflicting, li2025clone, ze2025twist, he2024omnih2o, he2024learning}. 
Existing research largely focuses on single-agent motion tracking \cite{he2025asap, zhang2025hub, xie2025kungfubot}, progressing from mimicking specific references to general motion tracking frameworks \cite{chen2025gmt, han2025kungfubot2, zhang2025track, ze2025twist, liao2025beyondmimic}. 
Beyond isolation, recent works have addressed the interaction between humanoids and environmental factors in sports such as table tennis \cite{su2025hitter}, badminton \cite{liu2025humanoid}, and soccer \cite{xu2025learning, ren2025humanoid}, as well as human-robot collaboration \cite{du2025learning, huang2026learning, chen2025symbridge}. 
However, these scenarios typically model the partner (human or object) as a passive entity or an external disturbance, ignoring the complex coupled dynamics inherent in multi-agent systems. Achieving interactive whole-body control on dual-humanoid hardware requires moving beyond these assumptions to explicitly model the mutual physical influence and geometric topology between active agents—a capability that remains absent in current robotics literature.

\subsection{Physics-Based Animation of Human-Human Interaction} While robotics research focuses on hardware feasibility, the computer graphics community has expanded the scope of humanoid animation from single-agent generation \cite{luo2023perpetual,peng2018deepmimic,peng2022ase} to the complex domain of \textit{Human-Human Interaction} \cite{wang2026intermoe, xu2025perceiving, starke2020local}.
Addressing the inevitable physical artifacts caused by sensor inaccuracies in existing datasets \cite{yin2023hi4d,liang2024intergen,xu2024interx,ghosh2024remos}, recent works \cite{yao2025physiinter,won2021control,gao2024coohoi,zhang2023simulation} employ RL within physics-based simulations to guarantee human-like behavior and physical plausibility.
However, these methods prioritize visual fidelity over dynamic feasibility, often relaxing rigorous constraints essential for real robots. Policies trained in these idealized environments struggle with the Sim-to-Real gap.
To bridge this gap, we propose \textbf{Rhythm}, which, to our knowledge, represents the first demonstration of physically coupled dual-humanoid interaction on real-world hardware.

\subsection{Motion Retargeting for Humanoid}
Motion retargeting serves as a fundamental bridge for transferring human skills to humanoids. However, standard approaches like PHC \cite{luo2023perpetual} and GMR \cite{araujo2025retargeting} typically treat agents in isolation, neglecting interaction behaviors. 
While OmniRetarget \cite{yang2025omniretarget} advances this by employing \textit{Interaction Meshes} \cite{alexa2003differential, zhou2005large} to enforce human-object interaction constraints, its applicability is restricted to single-humanoid settings. 
Extending this to human-human interaction presents unique challenges. Recent attempts, such as PAIR \cite{huang2026learning} and Harmanoid \cite{liu2025takes}, adopt a coupled formulation that intertwines self-motion with interaction constraints. Consequently, in the presence of inherent kinematic conflicts, they often compromise self-motion fidelity (e.g., causing unnatural distortions) to strictly enforce interaction geometry.
To address this, we propose IAMR that effectively mitigates these kinematic conflicts to produce high-fidelity coupled motion references.

\begin{figure*}[t]
\centerline{\includegraphics[width=1.02\textwidth]{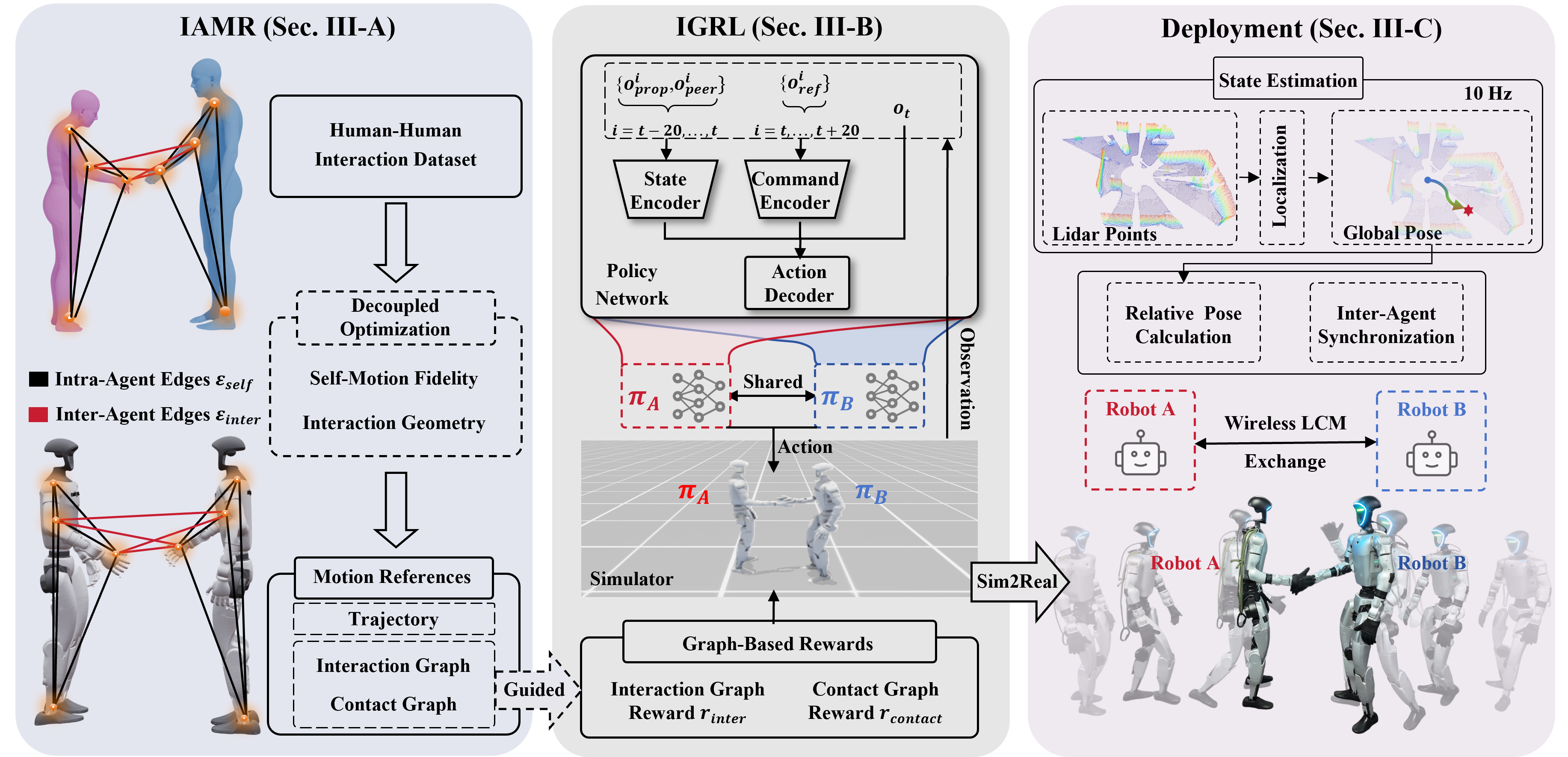}}
\vspace{-0.1cm} 
\caption{
\textbf{Overview of Rhythm.}
IAMR utilizes decoupled optimization to generate high-quality humanoid-humanoid motion interaction references from human demonstrations. Guided by these references, IGRL employs MAPPO and graph-based rewards to learn robust coupled dynamics. Finally, the deployment module facilitates Sim-to-Real transfer via Lidar-fused state estimation and inter-agent synchronization.
}
\label{fig:pipeline}
\vspace{-0.5cm}
\end{figure*}

\section{Methods}
\label{sec:methods}

As illustrated in Fig.~\ref{fig:pipeline}, \textbf{Rhythm} addresses the challenge of dual-humanoid interaction through three tightly integrated components: 
1)  \textbf{Interaction-Aware Motion Retargeting (IAMR)}, which synthesizes physically feasible priors by decoupling interaction geometry from self-motion fidelity (Sec.~\ref{subsec:retargeting}); 
2)  \textbf{Interaction-Guided Reinforcement Learning (IGRL)}, which captures coupled dynamics via topology-aware rewards that enforce interaction consistency (Sec.~\ref{subsec:rl}); 
and 3)  \textbf{Real-World Deployment}, which overcomes the limitations of noisy global observability and asynchronous execution to bridge the Sim-to-Real gap (Sec.~\ref{subsec:deployment}).

\subsection{Interaction-Aware Motion Retargeting (IAMR)}
\label{subsec:retargeting}

\subsubsection{Problem Formulation}
Given a source motion sequence of two human demonstrators with different anthropometry, our goal is to synthesize kinematically feasible trajectories for two humanoids. The core challenge lies in synthesizing motions that simultaneously preserve the individual motion style and the dense interaction geometry.

\noindent \textbf{Interaction Mesh and Laplacian Coordinates.}
To mathematically model the coupled multi-agent system, we adopt the volumetric Interaction Mesh formalism \cite{alexa2003differential, zhou2005large}. We represent the system as a connected graph $\mathcal{G} = (\mathcal{V}, \mathcal{E})$, comprising a vertex set $\mathcal{V}$ of the agents' key joints and an edge set $\mathcal{E}$ encoding their structural connections.
To encode local geometric details, we utilize Laplacian coordinates. For a vertex $p_i$, the Laplacian operator $\mathcal{L}(\cdot)$ computes the deviation from the weighted average of its neighbors $\mathcal{N}(i)$:
\begin{equation*}
  \mathcal{L}(p_i) = p_i - \sum_{j \in \mathcal{N}(i)} c_{ij} p_j,
  \label{eq:laplacian_def}
\end{equation*}
where $c_{ij}$ denotes the normalized weights.
In mesh-based motion retargeting, the objective is to find a target configuration $q$ such that the local geometry of the retargeted vertices $p_i(q)$ matches the source reference. This is achieved by minimizing the deformation energy $\sum \| \mathcal{L}(p_i(q)) - \mathcal{L}(p_i^{src}) \|^2$.

\noindent \textbf{The Kinematic Conflict.}
While the standard formulation $\sum \| \mathcal{L}(p_i(q)) - \mathcal{L}(p_i^{src}) \|^2$ is effective for single-agent editing, applying it to the transfer from heterogeneous humans to homogeneous robots creates a fundamental ambiguity in defining the source reference $\mathcal{V}^{src}$.
Due to the embodiment mismatch, no single reference manifold can simultaneously satisfy both self-motion and interaction constraints:
\begin{itemize}
    \item \textbf{Individual Manifold ($\mathcal{M}_{ind}$):} Constructed by scaling each human with individual ratios to match the robot's height. Using this as $\mathcal{V}^{src}$ preserves valid self-motion Laplacian coordinates but structurally disrupts the relative interaction geometry (e.g., causing ``air handshakes'').
    \item \textbf{Unified Manifold ($\mathcal{M}_{uni}$):} Constructed by applying a single global scale to the entire scene. Using this as $\mathcal{V}^{src}$ strictly preserves relative interaction edges but forces the robots to adopt kinematic constraints incompatible with their morphology (e.g., foot floating).
\end{itemize}

\noindent \textbf{Topological Partitioning.}
To resolve this conflict, we propose to relax the monolithic structure of the standard Interaction Mesh. Instead of treating the system as a single deformable body, we explicitly partition $\mathcal{E}$ into two disjoint functional groups, as visualized in the left part of Fig.~\ref{fig:pipeline}: (1) \textbf{Intra-Agent Edges ($\mathcal{E}_{self}$):} Edges connecting joints within a single robot, encoding the local self-motion topology. (2) \textbf{Inter-Agent Edges ($\mathcal{E}_{inter}$):} Edges connecting key joints between the two robots, encoding the relative interaction topology.

This topological decomposition allows us to assign distinct geometric references to different semantic parts of the graph, forming the basis for our decoupled optimization scheme.

\subsubsection{Decoupled Optimization}
Leveraging the topological partitioning defined above, we resolve the kinematic conflict by assigning distinct geometric references to the partitioned subgraphs. 
We formulate the retargeting as a dynamic spring system, where a self-motion term $E_{self}$ ensures intra-agent edges $\mathcal{E}_{self}$ track the Independent Manifold $\mathcal{M}_{ind}$, while an interaction term $E_{inter}$ constrains inter-agent edges $\mathcal{E}_{inter}$ to the Unified Manifold $\mathcal{M}_{uni}$.

We solve for the optimal joint configuration $q^*=\{q_1, q_2\}$ by minimizing a hybrid energy function:
\begin{equation*}
    q^* = \arg \min_q \left( E_{self}(q) + E_{inter}(q) \right) \quad \text{s.t.} \quad q \in \mathcal{C}_{phy},
    \label{eq:total_optimization}
\end{equation*}
where $\mathcal{C}_{phy}$ represents the set of feasible configurations satisfying joint limits and collision constraints.

\noindent \textbf{Self-Motion Objective ($E_{self}$).} To preserve individual motion quality, we align the Laplacian geometry of each robot to its Independent Reference $\mathcal{M}_{ind}$, strictly confining the operator to the local subgraph:
\begin{equation*}
\begin{split}
    E_{self}(q) = & \sum_{a \in \{1,2\}} \sum_{p_i \in \mathcal{V}_a} \| \mathcal{L}(p_i) - \mathcal{L}(p_i^{ind}) \|^2 \\
    & + \lambda_{rot} \sum_{a \in \{1,2\}} \sum_{k \in \mathcal{B}_{a}} \| \theta_k \ominus \hat{\theta}_k^{src} \|^2,
\end{split}
\label{eq:obj_self}
\end{equation*}
where $\mathcal{V}_a$ denotes the vertex set of robot $a$, and $p_i^{ind}$ represents the corresponding vertex position in the reference $\mathcal{M}_{ind}$. 
For rotation, $\mathcal{B}_{a}$ denotes the key links, where $\ominus$ measures the geodesic distance on $SO(3)$ between the current orientation $\theta_k$ and the reference $\hat{\theta}_k^{src}$.

\noindent \textbf{Interaction Objective ($E_{inter}$).}
To enforce the relative interaction geometry, we treat the inter-agent edges as extrinsic constraints driven by the Unified Reference $\mathcal{M}_{uni}$. We formulate this as a variable-stiffness spring potential:
\begin{equation*}
   E_{inter}(q) = \sum_{(i,j) \in \mathcal{E}_{inter}} \omega_{ij}(d_{ij}) \cdot \| (p_i - p_j) - (\hat{p}_i^{uni} - \hat{p}_j^{uni}) \|^2,
  \label{eq:obj_inter}
\end{equation*}
Here, $p_i$ and $p_j$ denote the current vertex positions of different agents, while $\hat{p}_i^{uni}$ and $\hat{p}_j^{uni}$ represent the corresponding target coordinates derived from the reference $\mathcal{M}_{uni}$. To naturally prioritize close-range geometry (e.g., contact) over distant relations, we define the stiffness $\omega_{ij}$ as a continuous exponential decay function of the source distance $d_{ij}$:
\begin{equation*}
    \omega_{ij}(d_{ij}) = \omega_{max} \cdot e^{-\gamma d_{ij}},
    \label{eq:dynamic_weight}
\end{equation*}
where $\omega_{max}$ denotes peak stiffness and $\gamma$ the decay rate.
This effectively models the interaction as a non-linear spring system that stiffens during close contact to prevent penetration, while becoming compliant at a distance to allow free motion.

\noindent \textbf{Topological Interaction Priors.}
Beyond kinematic trajectories, IAMR explicitly extracts inter-agent topological structures to serve as interaction priors for downstream policy learning, as illustrated in the left part of Fig.~\ref{fig:pipeline}.
We generate two cross-agent graph representations:
(1) An \textbf{Interaction Graph} (derived from $\mathcal{E}_{inter}$), which encodes the binary connectivity \textit{bridging} the keypoints of the two agents \cite{zhang2023simulation};
(2) A \textbf{Contact Graph} (constructed via collision detection), which records the binary physical contact states \textit{between} the links of the two distinct robots \cite{Wang_2025_CVPR}.
These graphs provide the essential interaction topology required for the graph-based rewards in the subsequent training phase.

\subsection{Interaction-Guided Reinforcement Learning (IGRL)}
\label{subsec:rl}

To master the coupled dynamics of dual-humanoid interaction, we propose \textbf{IGRL}, a multi-agent reinforcement learning module. Unlike standard motion tracking policies that treat agents as isolated entities, IGRL explicitly models interaction geometry through graph-based rewards and incorporates specific training strategies to bridge the reality gap.

\subsubsection{Multi-Agent Policy Design}
As shown in the middle part of Fig.~\ref{fig:pipeline}, we formulate the problem as a Multi-Agent Markov Decision Process (MA-MDP) and adopt the \textit{Centralized Training with Decentralized Execution} (CTDE) paradigm using MAPPO \cite{gao2024coohoi, littman1994markov}.

\vspace{2pt}
\noindent \textbf{Interaction-Centric Observation.}
Effective collaboration requires a comprehensive awareness of both self and partner states. The policy operates on a compact observation space $o_t = \{o_{prop}, o_{peer}, o_{ref}\}$:
\begin{itemize}
    \item Proprioception ($o_{prop}$): Defines the agent's internal state, including joint positions, velocities, base angular velocities, and the previous action.
    
    \item Peer Perception ($o_{peer}$): Encodes the peer's state relative to the ego agent. It includes the peer's joint positions and the relative root transform expressed in the ego-centric frame: the relative position $P_{rel} = R_{ego}^T (P_{peer} - P_{ego})$ and orientation $R_{rel} = R_{ego}^T R_{peer}$. This explicit spatial formation enables the agent to anticipate and handle coupled dynamics.
    
    \item Reference Motion ($o_{ref}$): Contains the future reference trajectories and the reference relative state, serving as the correct interaction topology.
\end{itemize}

\vspace{2pt}
\noindent \textbf{Network Architecture.}
We design a hierarchical policy architecture where 1D-CNN temporal encoders process historical observations and future references, feeding latent features along with the current observation into an MLP action decoder. 
More details are provided in Appendix B.

\vspace{2pt}
\noindent \textbf{Robust Training Strategy.}
To ensure transferability and handle complex interaction phases, we implement two strategies, with mathematical formulations provided in Appendix B:
\begin{itemize}
    \item \textbf{Curriculum-based Adaptive Sampling:} Standard RSI \cite{peng2018deepmimic} relies on sparse failure counts, neglecting non-terminal interaction violations. We propose an error-aware sampling strategy based on a multi-objective landscape composed of integrating failure, tracking, and interaction metrics. The curriculum dynamically evolves from prioritizing stability in early stages to focusing on tracking and interaction precision as the policy matures.
    \item \textbf{Dual-Agent Domain Randomization:} To bridge the sim-to-real gap, we simulate wireless latency via noisy, delayed peer observations and apply initial state perturbations to enforce recovery from physical misalignment.
\end{itemize}

\subsubsection{Graph-based Rewards}
Standard tracking rewards typically treat agents as isolated entities, failing to enforce topological consistency or capture coupled dynamics. 
To bridge this gap, we introduce graph-based rewards to \textbf{guide} the learning of dynamic control policies by translating the \textit{Topological Interaction Priors} established in IAMR, as shown in Fig.~\ref{fig:pipeline}.

\vspace{2pt}
\noindent \textbf{Interaction Graph Reward ($r_{inter}$).}
To ensure precise relative geometry during interaction, we penalize deviations of the \textit{interaction graph}. The reward is formulated as:
\begin{equation*}
r_{inter} = \exp\left( - \frac{1}{\sigma_{inter}} \sum_{(i,j) \in \mathcal{E}_{inter}} \omega_{ij} \| d_{ij}^{sim} - \hat{d}_{ij}^{ref} \|^2 \right),
\end{equation*}
where $\sigma_{inter}$ is a sensitivity scaling factor, and $d_{ij}$ denotes the relative position vector connecting joints $i$ and $j$ in the simulation ($sim$) and reference ($ref$) states.
By inheriting the distance-aware dynamic weights $\omega_{ij}$ from IAMR, the policy inherently learns to prioritize the same geometric constraints that were optimized during retargeting.

\vspace{2pt}
\noindent \textbf{Contact Graph Reward ($r_{contact}$).}
Since kinematic references lack force information, we utilize \textit{contact graph} to regularize physical interaction. This reward is designed with two goals: (1) \textit{Contact Consistency}, which penalizes mismatches between the simulated and reference contact states; and (2) \textit{Force Regularization}, which constrains contact forces to realistic ranges when active, and penalizes non-zero forces during non-contact phases. This explicitly discourages interpenetration and encourages compliant physical interaction.

\subsection{Real-World Deployment}
\label{subsec:deployment}

Deploying \textbf{Rhythm} on physical hardware faces the \textit{Sim-to-Real Gap}, specifically the lack of global observability and the issue of asynchronous execution.

\subsubsection{State Estimation and Relative Localization}

Constructing the observation space for dual-humanoid interaction requires precise global and relative state information. 
We employ a robust localization system, utilizing \textit{POINT-LIO} \cite{he2023point} for high-frequency local odometry and \textit{GICP} \cite{segal2009generalized} to register real-time point clouds against a pre-built map for drift-free global positioning. A Kalman Filter fuses these estimates to ensure robustness under highly dynamic motions.

Robots broadcast global poses $\{P, R\}$ via \textit{LCM} \cite{huang1998lcm}. The ego-agent transforms received data into its local frame to derive $\{P_{rel}, R_{rel}\}$, enabling real-time reconstruction of $o_{peer}$ consistent with the simulation.

\subsubsection{Inter-Agent Synchronization}

We use the \textit{motion phase} $\phi$ \cite{peng2018deepmimic} as the continuous variable representing the temporal execution progress of the interaction policy. In distributed settings, inherent hardware clock drift inevitably causes the phases of the two agents to diverge over time.

To maintain temporal alignment, we implement a soft synchronization mechanism based on proportional feedback. Agents exchange their current phase $\phi$ via the wireless bridge. Upon receiving the peer's phase $\phi_{peer}$, the ego agent dynamically modulates its phase progression rate $\dot{\phi}_{ego}$. Instead of a fixed increment ($\dot{\phi}_{base} = 1.0$), we apply a correction:
\begin{equation*}
\dot{\phi}_{ego} = 1.0 + k (\phi_{peer} - \phi_{ego}),
\end{equation*}
where $k$ is the synchronization gain. This compensates for temporal drift by modulating execution rate, ensuring smooth alignment without the discontinuities of hard phase resets.

\section{Experiments}
\label{sec:experiments}

We design our experiments to systematically validate the proposed framework by answering three core questions: 

\begin{itemize}
    \item \textbf{Q1 (Retargeting Quality):} Can IAMR synthesize kinematically feasible and topologically consistent trajectories that preserve the interaction geometry of the source data?
    \item \textbf{Q2 (Policy Efficacy):} Can IGRL utilize the interaction guidance to capture the coupled dynamics, overcoming the limitations of treating agents as isolated entities?
    \item \textbf{Q3 (Real-World Robustness):} Can the learned policies be successfully deployed to physical dual-humanoid systems, overcoming the limitations of noisy global observability  and asynchronous communication?
\end{itemize}

\subsection{The MAGIC Dataset}
\label{subsec:dataset}

\begin{figure}[t]
  \centering
  \includegraphics[width=\linewidth]{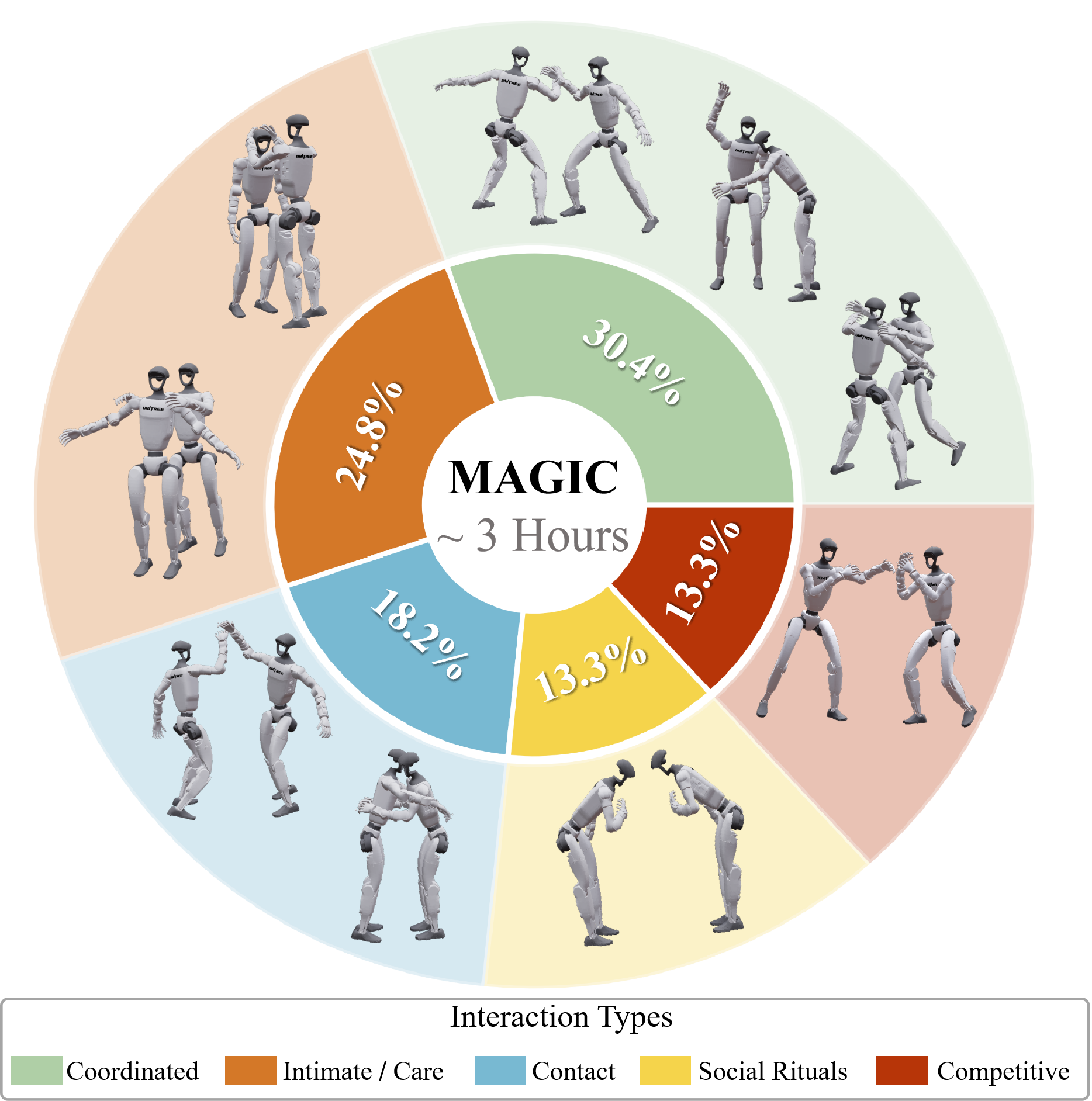}
  \vspace{-0.3cm}
  \caption{\textbf{Overview of MAGIC.} 
  MAGIC contains $\sim$3 hours of high-fidelity interaction data balanced across five semantic categories (inner chart). 
  Representative snapshots (outer ring) illustrate the diversity ranging from loose spatiotemporal coordination to intensive contact.}
  \label{fig:dataset_pie}
  \vspace{-0.5cm}
\end{figure}

High-quality motion data is the cornerstone of learning interactions. Addressing the lack of clean human-human interaction data, we introduce \textbf{MAGIC}, a high-fidelity motion capture dataset comprising $\sim$3 hours of valid motion sequences.

\subsubsection{Acquisition and Diversity}
Distinct from existing datasets \cite{yin2023hi4d,liang2024intergen,xu2024interx,ghosh2024remos}, we ensured physical plausibility for robot transfer through \textit{Anthropometric Consistency} (matched actor heights) and \textit{Temporal Continuity} (long-horizon sequences $>10$s). 
As illustrated in Fig.~\ref{fig:dataset_pie}, MAGIC covers a diverse semantic spectrum: \textit{Coordinated} actions (30.4\%), \textit{Intimate/Care} behaviors (24.8\%), \textit{Contact} (18.2\%), \textit{Social Rituals} (13.3\%), and \textit{Competitive} interactions (13.3\%).
All sequences are captured via high-fidelity optical motion capture in BVH format. 
Interaction graphs are extracted from the inter-agent edges $\mathcal{E}_{inter}$ 
of the retargeted motion, while contact labels are derived post-retargeting via 
collision detection on the retargeted humanoid kinematics.

\subsubsection{Data Release}
To facilitate future research, we will publicly release both the raw motion capture data and the retargeted humanoid reference trajectories generated by IAMR.

\subsection{Experimental Setup}
\label{subsec:setup}

\textbf{Datasets.}
We use the proposed MAGIC dataset for both training and evaluation. To systematically assess performance across diverse physical dynamics, we categorize the tasks into three physics-based groups: \textbf{Collaborate} (non-contact synchronization), \textbf{Light Contact} (transient interaction), and \textbf{Intensive Contact} (continuous force transmission). Additionally, we employ the external \textit{Inter-X} dataset~\cite{xu2024interx} to assess robustness under significant anthropometric mismatches.

\textbf{Baselines.}
We benchmark \textbf{Rhythm} against representative methods under two evaluation aspects, with full implementation details in Appendix C:
\begin{itemize}
    \item \textbf{Retargeting Quality (Q1):} We compare against \textbf{GMR}~\cite{araujo2025retargeting}, which performs Cartesian optimization, \textbf{OR}~\cite{yang2025omniretarget}, a single-agent interaction mesh method, and \textbf{DOR}, our constructed dual-agent extension of OR.

    \item \textbf{Policy Efficacy (Q2):} To validate the effectiveness of IGRL, we compare against a \textbf{Single-Agent} (Status Quo) baseline that performs isolated tracking, as well as key ablated variants of our method, including \textbf{w/o Peer Obs}, \textbf{w/o Contact Rew}, and \textbf{w/o Interaction Rew}.
\end{itemize}

\textbf{Evaluation Metrics.}
We adopt task-specific evaluation metrics for
\textbf{Retargeting Quality (Q1)} and \textbf{Policy Efficacy (Q2)}.
For conciseness, detailed metric definitions and mathematical
formulations are deferred to Appendix~C.

\subsection{Retargeting Quality (Q1)}
\label{subsec:retargeting_results}

We evaluate IAMR against baselines across three dimensions: 
\textbf{Safety}, measured by Inter-Penetration Rate (IPR) and Max Penetration Depth (MPD); 
\textbf{Fidelity}, quantified by Interaction Edge Error (IEE) and Contact F1 Score~\cite{liu2025takes,huang2026learning}; 
and \textbf{Utility}, assessed via Downstream Success Rate (DSR).

\begin{table}[t]
\centering
\caption{\textbf{Quantitative Results of Retargeting.} Comparison across four interaction categories. Metrics include Safety (IPR, MPD), Fidelity (IEE, F1), and Utility (DSR). IAMR achieves the best balance, strictly eliminating penetration (IPR=0) while maximizing contact F1 scores.}
\label{tab:retarget_quantitative}

\definecolor{Gray}{gray}{0.9}

\small 
\renewcommand{\arraystretch}{1.15}
\setlength{\tabcolsep}{0pt}

\begin{tabular*}{\columnwidth}{@{\extracolsep{\fill}} @{\hspace{8pt}} l cccccc @{\hspace{8pt}} }
\toprule
\multirow{2}{*}{} & \multicolumn{2}{c}{\textbf{Safety}} & \multicolumn{3}{c}{\textbf{Fidelity}} & \textbf{Utility} \\
\cmidrule(lr){2-3} \cmidrule(lr){4-6} \cmidrule(lr){7-7}
 & IPR\scriptsize{(\%)$\downarrow$} & MPD\scriptsize{(cm)$\downarrow$} & IEE\scriptsize{(\%)$\downarrow$} & F1-S\scriptsize{$\uparrow$} & F1-L\scriptsize{$\uparrow$} & DSR\scriptsize{(\%)$\uparrow$} \\
\midrule

\rowcolor{Gray} \multicolumn{7}{l}{\textbf{\textit{MAGIC: Collaborate}}} \\
\hspace{1em}GMR & \underline{0.14} & \underline{1.2} & 4.3 & 0.602 & 0.804 & 85.5 \\
\hspace{1em}OR  & 0.20 & 1.4 & 4.1 & \underline{0.747} & \underline{0.902} & 87.4 \\
\hspace{1em}DOR & \textbf{0.00} & \textbf{0.0} & \underline{3.9} & 0.711 & 0.899 & \textbf{89.5} \\
\hspace{1em}\textbf{IAMR} & \textbf{0.00} & \textbf{0.0} & \textbf{3.7} & \textbf{0.785} & \textbf{0.936} & \underline{89.0} \\

\rowcolor{Gray} \multicolumn{7}{l}{\textbf{\textit{MAGIC: Light Contact}}} \\
\hspace{1em}GMR & \underline{2.18} & \underline{3.3} & 4.6 & 0.738 & 0.893 & 48.5 \\
\hspace{1em}OR  & 7.62 & 5.9 & \underline{3.6} & \underline{0.844} & 0.912 & 63.1 \\
\hspace{1em}DOR & \textbf{0.00} & \textbf{0.0} & \underline{3.6} & 0.810 & \underline{0.918} & \underline{69.4} \\
\hspace{1em}\textbf{IAMR} & \textbf{0.00} & \textbf{0.0} & \textbf{3.1} & \textbf{0.905} & \textbf{0.935} & \textbf{75.3} \\

\rowcolor{Gray} \multicolumn{7}{l}{\textbf{\textit{MAGIC: Intensive Contact}}} \\
\hspace{1em}GMR & \underline{35.2} & \underline{3.8} & 9.6 & 0.864 & 0.928 & 45.5 \\
\hspace{1em}OR  & 47.3 & 5.3 & 8.0 & \underline{0.884} & \underline{0.929} & 56.5 \\
\hspace{1em}DOR & \textbf{0.00} & \textbf{0.0} & \underline{7.8} & 0.883 & 0.925 & \underline{63.3} \\
\hspace{1em}\textbf{IAMR} & \textbf{0.00} & \textbf{0.0} & \textbf{6.6} & \textbf{0.932} & \textbf{0.941} & \textbf{78.3} \\

\rowcolor{Gray} \multicolumn{7}{l}{\textbf{\textit{Inter-X}}} \\
\hspace{1em}GMR & \underline{11.7} & \underline{1.7} & 8.0 & \underline{0.598} & 0.752 & 31.7 \\
\hspace{1em}OR  & 18.4 & 2.6 & 6.8 & 0.587 & 0.791 & 46.3 \\
\hspace{1em}DOR & \textbf{0.00} & \textbf{0.0} & \underline{6.7} & 0.589 & \underline{0.795} & \underline{52.9} \\
\hspace{1em}\textbf{IAMR} & \textbf{0.00} & \textbf{0.0} & \textbf{4.9} & \textbf{0.843} & \textbf{0.860} & \textbf{69.9} \\

\bottomrule
\end{tabular*}
\vspace{-0.5cm}
\end{table}

\begin{figure}[t!]
  \centering
  \includegraphics[width=\linewidth]{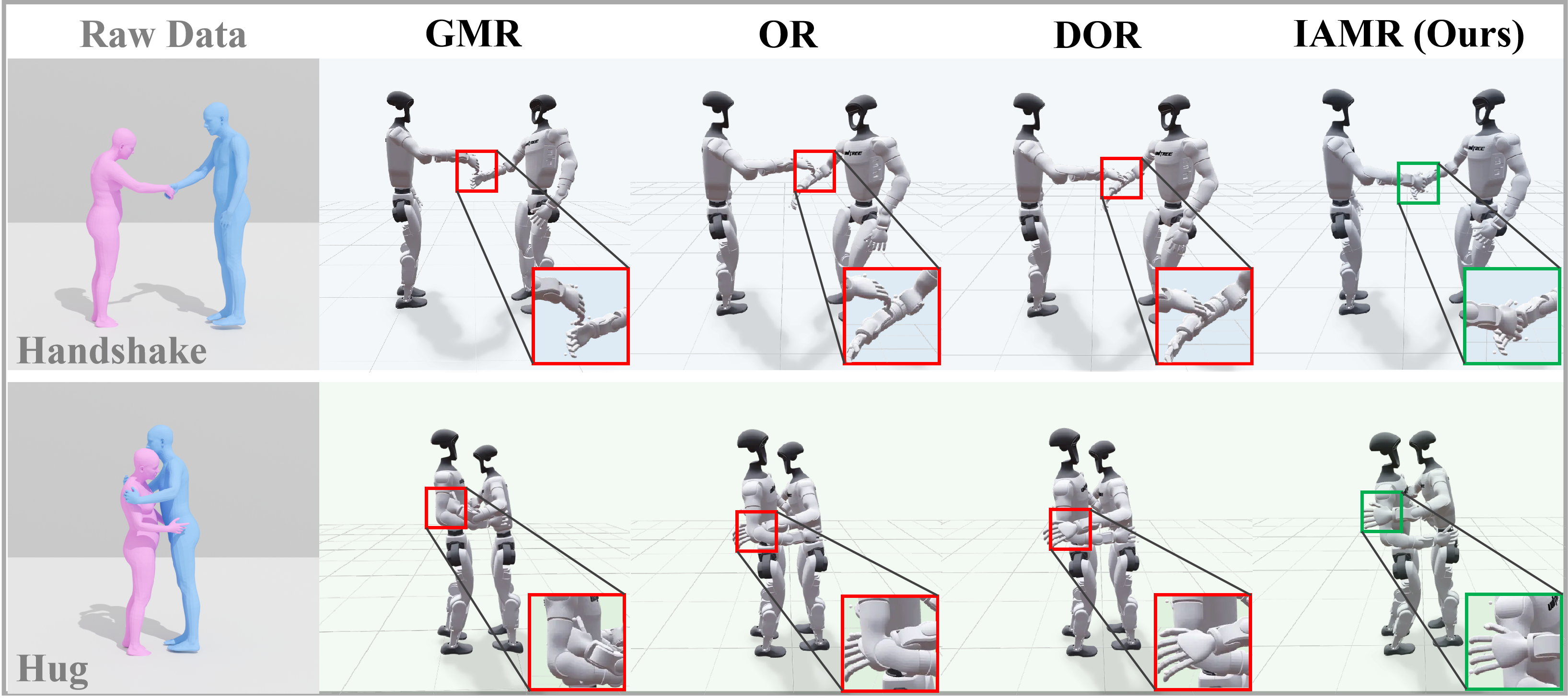} 
  \vspace{-0.4cm}
  \caption{\textbf{Qualitative Visualization of Retargeting on Inter-X.} 
  \textbf{Top:} Baselines suffer from contact loss (``air handshakes''), whereas IAMR preserves precise interaction geometry. 
  \textbf{Bottom:} OR leads to severe penetration while DOR forces unnatural stiff postures; IAMR maintains close-proximity topology without collisions.}
  \label{fig:retarget_demo}
    \vspace{-0.3cm}
\end{figure}

\textbf{Quantitative Analysis.} 
Table~\ref{tab:retarget_quantitative} presents the performance comparison across four interaction categories. We observe three key trends. 
First, isolated baselines (GMR, OR) fundamentally fail to ensure physical feasibility. Notably, in Intensive Contact, OR exhibits a prohibitive Inter-Penetration Rate (IPR) of 47.3\%, rendering the motions physically infeasible and unsuitable for policy learning. 
Second, while the coupled baseline (DOR) guarantees safety (IPR=0), its rigid formulation compromises interaction fidelity under anthropometric mismatch. This is evident on the Inter-X dataset, where DOR's F1-Strict drops to 0.589 due to its inability to reconcile conflicting kinematic constraints.
In contrast, IAMR achieves the optimal safety-fidelity trade-off. By adaptively decoupling self-motion from interaction objectives, our method eliminates penetration while outperforming DOR by 43\% in F1-Strict on Inter-X. Crucially, this strict preservation of interaction topology translates to superior downstream utility, yielding 78.3\% DSR in contact-rich tasks.

\begin{table}[t!]
\centering
\caption{\textbf{Quantitative Results of Policy.} We evaluate the contribution of each component. Our full method achieves the most robust balance, effectively integrating coarse-grained geometric alignment (low IEE) with fine-grained physical contact fidelity (high CSR).}
\label{tab:policy_ablation}

\definecolor{Gray}{gray}{0.9}

\small 

\renewcommand{\arraystretch}{1.15}
\setlength{\tabcolsep}{0pt}

\begin{tabular*}{\columnwidth}{@{\extracolsep{\fill}} @{\hspace{8pt}} l cccc @{\hspace{8pt}} }
\toprule
\multirow{2}{*}{} & \multicolumn{2}{c}{\textbf{Interaction}} & \multicolumn{2}{c}{\textbf{Contact}} \\
\cmidrule(lr){2-3} \cmidrule(lr){4-5}
 & ISR\scriptsize{(\%)$\uparrow$} & IEE\scriptsize{(\%)$\downarrow$} & CSR\scriptsize{(\%)$\uparrow$} & CER\scriptsize{$\downarrow$} \\
\midrule

\rowcolor{Gray} \multicolumn{5}{l}{\textbf{\textit{MAGIC: Collaborate}}} \\
\hspace{1em}Single Agent      & 18.7 & 38.9 & 100.0 & 0.000 \\
\hspace{1em}w/o Peer Obs      & 19.5 & 47.0 & 100.0 & 0.000 \\
\hspace{1em}w/o Contact Rew   & \textbf{93.4} & \textbf{4.7}  & 100.0 & 0.000 \\
\hspace{1em}w/o Interact Rew  & 58.1 & 15.1 & 100.0 & 0.000 \\
\hspace{1em}\textbf{Ours (Full)}       & \underline{92.9} & \underline{4.8}  & 100.0 & 0.000 \\

\rowcolor{Gray} \multicolumn{5}{l}{\textbf{\textit{MAGIC: Light Contact}}} \\
\hspace{1em}Single Agent      & 34.3 & 19.9 & 24.1 & 0.283 \\
\hspace{1em}w/o Peer Obs      & 48.9 & 13.9 & 18.6 & 0.268 \\
\hspace{1em}w/o Contact Rew   & \underline{85.9} & \underline{5.4}  & \underline{52.1} & \underline{0.203} \\
\hspace{1em}w/o Interact Rew  & 48.7 & 19.3 & 28.1 & 0.243 \\
\hspace{1em}\textbf{Ours (Full)}       & \textbf{90.0} & \textbf{4.2}  & \textbf{78.0} & \textbf{0.120} \\

\rowcolor{Gray} \multicolumn{5}{l}{\textbf{\textit{MAGIC: Intensive Contact}}} \\
\hspace{1em}Single Agent      & 24.0 & 29.9 & 37.5 & 0.312 \\
\hspace{1em}w/o Peer Obs      & 34.1 & 21.4 & 43.7 & 0.280 \\
\hspace{1em}w/o Contact Rew   & \textbf{77.3} & \textbf{7.7}  & \underline{70.6} & \underline{0.174} \\
\hspace{1em}w/o Interact Rew  & 51.3 & 17.0 & 56.8 & 0.211 \\
\hspace{1em}\textbf{Ours (Full)}       & \underline{75.2} & \underline{7.9}  & \textbf{78.8} & \textbf{0.159} \\

\rowcolor{Gray} \multicolumn{5}{l}{\textbf{\textit{Inter-X}}} \\
\hspace{1em}Single Agent      & 25.7 & 26.9 & 57.4 & 0.256 \\
\hspace{1em}w/o Peer Obs      & 73.2 & 7.6  & 75.3 & 0.143 \\
\hspace{1em}w/o Contact Rew   & \textbf{95.1} & \textbf{3.4}  & 68.3 & 0.208 \\
\hspace{1em}w/o Interact Rew  & 63.2 & 9.7  & \textbf{78.6} & \textbf{0.110} \\
\hspace{1em}\textbf{Ours (Full)}       & \underline{92.8} & \underline{3.5}  & \underline{77.4} & \underline{0.125} \\

\bottomrule
\end{tabular*}
\vspace{-0.5cm}
\end{table}

\begin{figure}[t]
  \centering
  \includegraphics[width=1\linewidth]{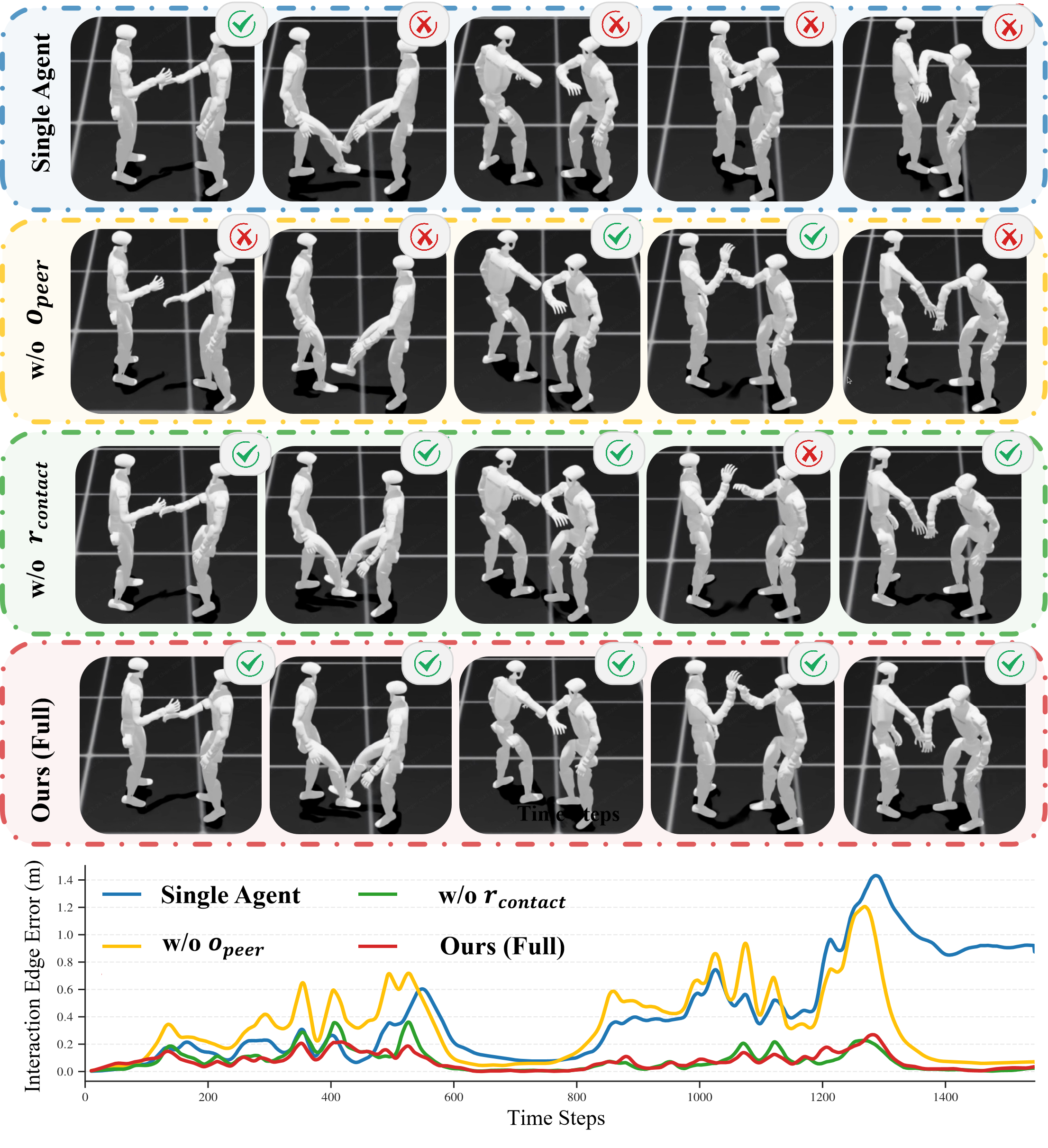}
  \vspace{-0.4cm}
  \caption{\textbf{Qualitative Visualization of Policy.} 
  Single Agent (blue) drifts into collisions. 
  w/o Contact Rew (green) achieves low error but exhibits physical ``ghosting''. 
  In contrast, Ours enforces valid physical contact.}
  \label{fig:policy_demo}
  \vspace{-0.3cm}
\end{figure}

\textbf{Qualitative Visualization (Inter-X Cases).} 
Fig.~\ref{fig:retarget_demo} validates robustness under significant mismatches. In transient ``Handshakes'', IAMR utilizes dynamic weighting to prioritize critical interaction geometry, effectively resolving the contact loss (``air handshakes'') observed in baselines. Conversely, in continuous ``Hugs'', our decoupled optimization reconciles kinematic conflicts, preventing the penetration of OR and the stiffness of DOR to maintain valid close-proximity topology.

\subsection{Policy Efficacy (Q2)}
\label{subsec:policy_results}

We assess the robustness and fidelity of the learned control policy across two dimensions: 
\textbf{Interaction Performance}, measured by Interaction Edge Error (IEE) and Interaction Success Rate (ISR); 
and \textbf{Contact Performance}, quantified by Contact Success Rate (CSR) and Contact Error Rate (CER).

\textbf{Quantitative Analysis.}
Table~\ref{tab:policy_ablation} presents the quantitative ablation results, revealing two critical insights. 
First, Interaction Awareness is non-negotiable. The Single Agent (Vanilla) baseline, limiting agents to isolated tracking, fails to coordinate effectively, achieving only 18.7\% ISR in Collaborate scenarios. Similarly, removing peer observations (w/o Peer Obs) severs the closed-loop synchronization, causing physical coupling to collapse (CSR drops to 18.6\% in Light Contact).
Second, there exists a functional hierarchy between coarse-grained geometric guidance and fine-grained contact regulation. 
The ablation results reveal that the interaction reward serves as a necessary foundation: removing it (w/o Interaction Rew) causes performance to collapse across all metrics as the policy fails to guide agents into the interaction envelope where contact can be established. For example, ISR drops to 58.1\% in Collaborate and CSR drops to 28.1\% in Light Contact.
Once this spatial proximity is achieved, the contact reward becomes critical for physical realism. 
Notably, the w/o Contact Rew variant exhibits a ``ghosting'' phenomenon: it attains high geometric precision (93.4\% ISR in Collaborate) by disregarding physical collision constraints, yet fails to maintain valid physical contact (only 52.1\% CSR in Light Contact). This is because the policy over-optimizes $r_{inter}$, exploiting physical interpenetration as a shortcut to minimize geometric error 
rather than establishing true physical coupling.
Ours (Full) effectively integrates these components, leveraging geometric guidance to establish the spatial foundation while using contact regulation to enforce valid physical coupling (above 75\% ISR and 77\% CSR across all scenarios).

\textbf{Qualitative Visualization.} 
Fig.~\ref{fig:policy_demo} highlights the behavioral divergence across methods in the Greeting task.
The Single Agent, treating the peer merely as a dynamic obstacle, may initiate interaction but inevitably drifts into collision. While its low-level robustness prevents immediate termination, the interaction topology is destroyed. 
Similarly, without explicit peer monitoring, the w/o Peer Obs baseline suffers from severe desynchronization (large error spikes). 
Notably, the w/o Contact Rew variant successfully maintains coarse-grained geometric alignment, yielding an IEE curve comparable to the full method (Green versus Red lines). However, it lacks fine-grained contact fidelity, allowing hands to ``ghost'' through each other. 
Ours bridges this gap, achieving the precision as w/o Contact Rew while enforcing valid physical coupling at the contact interface.
These visual observations directly align with the quantitative hierarchy in Table~\ref{tab:policy_ablation}, particularly explaining the discrepancy between geometric (ISR) and physical (CSR) metrics in the baselines.

\subsection{Real-World Robustness (Q3)}
We validate the deployment of \textbf{Rhythm} on Unitree G1 humanoids to assess its generality, robustness, and success rate in physical environments.

\noindent \textbf{Framework Generality.} 
Fig.~\ref{fig:teaser} demonstrates the system's versatility across diverse modalities. By strictly preserving fine-grained contact geometry in physical coupling tasks (Fig.~\ref{fig:teaser}a-c) and maintaining spatiotemporal coherence in long-horizon coordination (Fig.~\ref{fig:teaser}d), \textbf{Rhythm} effectively ensures interaction integrity across the spectrum using a unified formulation.

\begin{table}[t]
\centering
\caption{\textbf{Main Results for Real Robot Experiments.} We conducted 10 trials for each task and evaluated success based on contact establishment at specific keyframes ($K$ frames per trial).}
\label{tab:real_world_success}

\renewcommand{\arraystretch}{1.1}

\begin{tabular*}{\columnwidth}{@{\extracolsep{\fill}}llcc}
\toprule
\textbf{Task} & \textbf{Method} & \textbf{Success / Total} & \textbf{Rate (\%)} \\
\midrule

\multirow{2}{*}{\textbf{Hug}} & Single Agent & $8~/~30$ & 26.7\% \\
 & \textbf{Ours} & \textbf{26 / 30} & \textbf{86.7\%} \\
\midrule

\multirow{2}{*}{\textbf{Shoulder}} & Single Agent & $6~/~30$ & 20.0\% \\
 & \textbf{Ours} & \textbf{24 / 30} & \textbf{80.0\%} \\
\midrule

\multirow{2}{*}{\textbf{Greeting}} & Single Agent & $11~/~90$ & 12.2\% \\
 & \textbf{Ours} & \textbf{74 / 90} & \textbf{82.2\%} \\
 
\bottomrule
\end{tabular*}
\vspace{-0.3cm}
\end{table}

\begin{figure}[t]
    \centering
    \includegraphics[width=\linewidth]{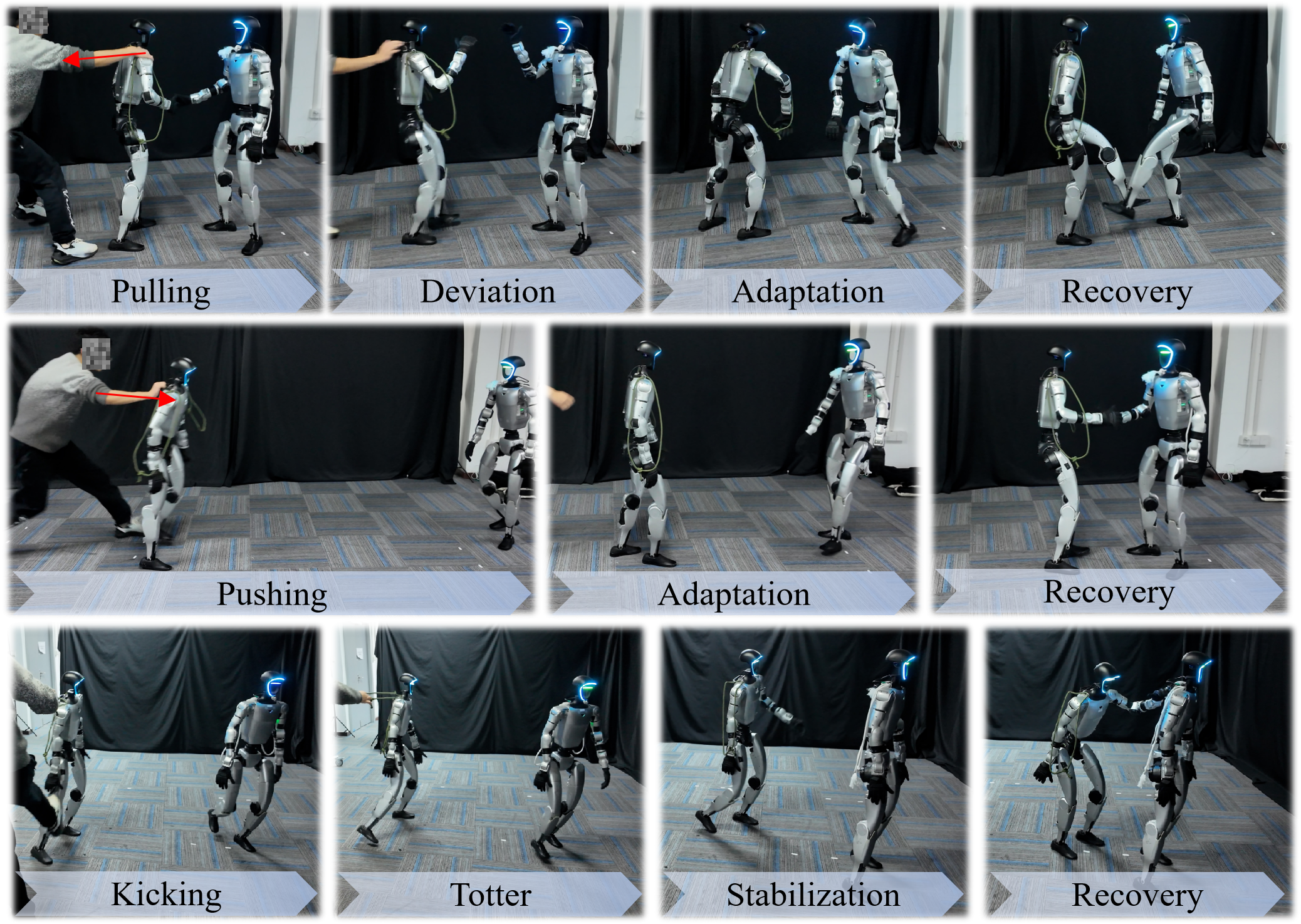} 
    \vspace{-0.1cm}
    \caption{\textbf{Robustness to disturbances.} Our policy demonstrates strong resilience against aggressive external perturbations (pulling, pushing, and kicking), successfully recovering balance and synchronization.}
    \label{fig:robustness}
    \vspace{-0.4cm}
\end{figure}

\noindent \textbf{Quantitative Success Rate.}
We evaluate success based on valid contact establishment at task-specific keyframes: \textit{Hug} uses 3 keyframes (shoulder pat, hand clasp, full hug); \textit{Shoulder} uses 3 keyframes (start, mid, end of walking); and \textit{Greeting} uses 9 keyframes (spanning hand, leg, shoulder, and elbow contacts).
Table~\ref{tab:real_world_success} reports success rates (measured over 10 trials). \textbf{Rhythm} consistently outperforms the Single Agent baseline by more than 60\%, with the most pronounced gap in \textit{Greeting} ($12.2\%$ versus $82.2\%$). While the baseline fails due to accumulated drift in this long-horizon task, our relative state estimation continuously compensates for misalignment, ensuring robust physical coupling.

\noindent \textbf{Robustness to Disturbances.}
To evaluate system resilience, we subjected the robots to significant external perturbations during execution. As illustrated in Fig.~\ref{fig:robustness}, these disturbances included aggressive pushing, pulling, and kicking forces. Despite the severity of these physical interferences, the agents successfully recovered their balance and actively adjusted their relative states to restore the interaction topology. This empirical evidence confirms that our policy has learned robust closed-loop synchronization capabilities, allowing for real-time recovery strategies rather than open-loop motion replay.

\section{Conclusion}
\label{sec:conclusion}
In this work, we present \textbf{Rhythm}, a unified framework that, to our knowledge, achieves the first robust transfer of physically coupled interactive behaviors to dual-humanoid hardware. By integrating Interaction-Aware Motion Retargeting (IAMR) to resolve kinematic conflicts and Interaction-Guided Reinforcement Learning (IGRL) to master coupled dynamics, our approach effectively bridges the Sim-to-Real gap. Furthermore, we release the MAGIC dataset to facilitate future research in multi-agent embodied intelligence.

While this work utilizes dual-humanoid setups to rigorously validate the modeling of coupled dynamics, our graph-based formulation is theoretically generic and supports extension to multi-agent systems. A current limitation is the reliance on pre-built maps for state estimation. Future work will focus on eliminating this dependency by shifting towards fully ego-centric perception for map-free collaboration, and scaling the framework to orchestrate complex multi-humanoid interactions in open-world scenarios.

\bibliographystyle{plainnat}
\bibliography{references}

\clearpage
\section*{Appendix}

\subsection*{Overview}
\label{sec:append_overview}
This appendix is organized into three main sections (A--C) to support the clarity and reproducibility of the proposed framework, \textbf{Rhythm}.

\vspace{2pt}
\noindent \textbf{Terminology.}
We refer to our unified framework as \textbf{Rhythm}. Within this framework, we define two core components:
\begin{itemize}
    \item \textbf{IAMR} (Interaction-Aware Motion Retargeting): The retargeting module that resolves kinematic conflicts to generate geometrically consistent motion references from heterogeneous human data (Sec. A).
    \item \textbf{IGRL} (Interaction-Guided Reinforcement Learning): The multi-agent learning module that masters coupled dynamics via graph-based rewards (Sec. B).
\end{itemize}

\vspace{2pt}
\noindent \textbf{Structure.}
The Appendix is organized as follows:
\begin{itemize}
    \item \textbf{A. Details of IAMR (Sec. A):} We first elaborate on the data processing pipeline for heterogeneous motion sources (Part 1). We then provide the mathematical formulations and constraints for the optimization problem (Part 2), followed by a visualization of the topological interaction priors (Part 3).
    
    \item \textbf{B. Details of IGRL (Sec. B):} This section specifies the hierarchical network architecture (Part 1), presents the comprehensive definition of graph-based rewards (Part 2), and details the robust training strategies, including curriculum learning and domain randomization (Part 3).
    
    \item \textbf{C. Experimental Setup \& Metrics (Sec. C):} We provide implementation details for the benchmarking baselines (Part 1), followed by the rigorous mathematical definitions of the evaluation metrics (Part 2). Finally, we describe the hardware configuration and localization system used for Sim-to-Real transfer (Part 3).
\end{itemize}

\subsection{Details of IAMR}
\label{sec:append_iamr}

\subsubsection{Compatibility with Heterogeneous Motion Sources}
\label{subsubsec:iamr_sources}

Human motion datasets contain rich pose and interaction information but differ significantly in data format and physical attributes (e.g., height, body proportions). A key strength of our framework is its input-agnostic design: we first abstract diverse inputs into a standardized representation—time-series of \textit{raw} global 3D keypoint positions $\{p^\text{raw}_{t, i}\}$—and then process them into two distinct reference manifolds to resolve the kinematic conflict.

\vspace{5pt}
\noindent \textbf{Standardization of Input Formats.}
Regardless of the source format, our first step is to extract the raw 3D keypoints $p^\text{raw}_{t, i}$ for each agent $k \in \{1, 2\}$:
\begin{itemize}
    \item \textbf{Parametric Human Models (SMPL):} The \textbf{Inter-X} dataset \cite{xu2024interx} utilizes the SMPL format. We compute the raw keypoints via the SMPL forward kinematics function $M(\cdot)$ using the recorded pose $q$ and shape $\beta$: 
    \begin{equation*}
        p^{\text{raw}, (k)}_{t} = M(q^{(k)}_t; \beta^{(k)}).
    \end{equation*}
    \item \textbf{Skeleton Hierarchy (BVH):} Our \textbf{MAGIC} dataset utilizes the standard BVH skeleton hierarchy. Here, raw keypoints are derived directly from the skeleton's forward kinematics $f^\text{skel}(\cdot)$:
    \begin{equation*}
        p^{\text{raw}, (k)}_{t} = f^\text{skel}(q^{(k)}_t).
    \end{equation*}
\end{itemize}

\vspace{5pt}
\noindent \textbf{Construction of Dual Reference Manifolds.}
As discussed in the \textit{Kinematic Conflict} (Sec.~\ref{subsec:retargeting}), directly using raw human keypoints is infeasible due to morphological mismatches. To decouple interaction semantics from individual embodiment, we generate two distinct sets of scaled reference keypoints, $P_{ind}$ and $P_{uni}$, corresponding to the individual ($\mathcal{M}_{ind}$) and unified ($\mathcal{M}_{uni}$) manifolds respectively.

Let $h_{\text{robot}}$ be the robot's height, and $h_{\text{raw}}^{(k)}$ be the height of the $k$-th human demonstrator derived from $p^{\text{raw}, (k)}$. We first compute the individual height ratio $s^{(k)} = h_{\text{robot}} / h_{\text{raw}}^{(k)}$.

\begin{itemize}
    \vspace{3pt}
    \item \textbf{Individual Scaling for Self-Motion ($\mathcal{M}_{ind}$):}
    To ensure kinematic feasibility for each robot's self-motion (e.g., limb proportions, ground contact), we generate the individual reference set $P_{ind}$ by scaling each agent's raw keypoints with its own specific ratio $s^{(k)}$:
    \begin{equation*}
        p^{\text{ind}, (k)}_{t, i} = s^{(k)} \cdot p^{\text{raw}, (k)}_{t, i}.
    \end{equation*}
    This set $P_{ind}$ serves as the target for all single-agent tracking objectives (e.g., $\mathcal{L}_{pos}$, $\mathcal{L}_{reg}$), ensuring that each robot tracks a trajectory compatible with its own scale.

    \vspace{3pt}
    \item \textbf{Unified Scaling for Interaction Geometry ($\mathcal{M}_{uni}$):}
    To preserve the relative interaction topology (e.g., hand-holding distance), we generate the unified reference set $P_{uni}$ by applying a single global scale $s_{unified}$ to both agents. We define this scale as the average of the individual factors:
    \begin{equation*}
        s_{unified} = \frac{s^{(1)} + s^{(2)}}{2}, \quad p^{\text{uni}, (k)}_{t, i} = s_{unified} \cdot p^{\text{raw}, (k)}_{t, i}.
    \end{equation*}
    This set $P_{uni}$ is exclusively used to compute the \textit{Interaction Graph} targets (e.g., relative edge lengths). This ensures that the spatial relationship between agents remains consistent with the original performance, preventing the geometric distortion that would arise from non-uniform scaling.
\end{itemize}

\subsubsection{Optimization Details and Hyperparameters}
\label{subsubsec:iamr_optim}

We explicitly formulate the optimization objectives and constraints used to solve the kinematic conflict described in Sec.~\ref{subsec:retargeting}.

\vspace{3pt}
\noindent \textbf{Optimization Formulation.}
We solve the retargeting problem frame-by-frame using a Sequential Quadratic Programming (SQP) approach. For each frame $t$, we optimize the joint configurations $q_t = [q_t^{(1)}, q_t^{(2)}]$ to minimize the following total objective:
\begin{equation*}
    \mathcal{J}(q_t) = w_{self} \mathcal{J}_{self} + w_{inter} \mathcal{J}_{inter} + w_{reg} \mathcal{J}_{reg}.
\end{equation*}

\noindent The individual terms are defined as follows:

\begin{itemize}
    \item \textbf{Self-Motion Preservation ($\mathcal{J}_{self}$):} 
    To maintain local topology and orientation, we minimize deviations from the \textit{individual} reference manifold $P_{ind}$.
    \begin{equation*}
    \small
    \begin{split}
        \mathcal{J}_{self} = \sum_{k \in \{1, 2\}} \Big( & \| \mathcal{L}(f(q_t^{(k)})) - \mathcal{L}(P_{ind}^{(k)}) \|^2 \\
        & + \lambda_{rot} \sum_{b \in \mathcal{B}_k} \| \theta_b(q_t^{(k)}) \ominus \hat{\theta}_b^{src} \|^2 \Big).
    \end{split}
    \end{equation*}
    Here, $\mathcal{L}(\cdot)$ denotes the discrete Laplacian operator, and $\theta_b$ represents key bone orientations. This term allows the robot to adapt its absolute posture while preserving local motion semantics.

    \item \textbf{Interaction Preservation ($\mathcal{J}_{inter}$):} 
    This term enforces relative geometric consistency by tracking the \textit{unified} reference manifold $P_{uni}$. We split the interaction error term to fit the column width:
    \begin{equation*}
    \small
    \begin{split}
        \mathcal{J}_{inter} = \sum_{(i, j) \in \mathcal{E}_{inter}} \omega_{ij} \big\| & (f_i(q_t^{(1)}) - f_j(q_t^{(2)})) \\
        & - (p_{t,i}^{uni, (1)} - p_{t,j}^{uni, (2)}) \big\|^2.
    \end{split}
    \end{equation*}
    Where $\omega_{ij}$ is the distance-dependent stiffness. This explicitly penalizes deviations in the relative position vectors between agents.

    \item \textbf{Regularization ($\mathcal{J}_{reg}$):} 
    Ensures temporal smoothness. It includes a minimum velocity term $\|q_t - q_{t-1}\|^2$.
\end{itemize}

\vspace{2pt}
\noindent \textbf{Constraints.} The optimization is subject to the following hard constraints:
\begin{itemize}
    \item \textbf{Joint Limits:} The solution must respect the robot's physical joint ranges: $q^{min} \le q_{t-1} + \Delta q_t \le q^{max}$.
    
    \item \textbf{Collision Avoidance:} We enforce non-penetration for all active collision pairs. Linearizing the signed distance field $\phi(q)$, we require $J_{col} \Delta q_t \ge -\phi(q_{t-1}) - \epsilon_{safe}$, where $J_{col}$ is the normal Jacobian and $\epsilon_{safe}$ is a safety margin.

    \item \textbf{Foot Contact:} For feet in strict contact (detected from source motion), we impose a zero-velocity constraint on the end-effectors: $\| J_{foot} \Delta q_t \| \le \epsilon_{stick}$.

    \item \textbf{Trust Region:} To ensure the validity of the linearization, we bound the step size: $\|\Delta q_t\|_2 \le \delta$.
\end{itemize}

\vspace{5pt}
\noindent \textbf{Implementation \& Hyperparameters.}
The optimization is implemented in Python using \textbf{CVXPY} with the \textbf{OSQP} solver. 
Empirically, we set the weights to prioritizes interaction stability: $w_{self}=2.0$, $w_{inter}=10.0$, and $w_{reg}=0.1$. The rotation weight $\lambda_{rot}$ is set to $0.1$.

\begin{algorithm}[t]
\small
\caption{Interaction-Aware Motion Retargeting (IAMR)}
\label{alg:iamr}
\begin{algorithmic}[1]
\Require Source Motion $Q^{src}$, Robot Models $\mathcal{R}_1, \mathcal{R}_2$
\Ensure Retargeted Motion $Q^{rob}$

\State \textbf{Phase 1: Dual-Manifold Construction}
\State Compute scales: $s_k \leftarrow h_{rob}/h_{src}^{(k)}$, $s_{uni} \leftarrow \text{avg}(s_k)$
\For{$t = 0 \to T$}
    \State $p^{raw}_t \leftarrow \text{FK}(q^{src}_t)$
    \State $P_{ind}^{(k)} \leftarrow s_k \cdot p^{raw, (k)}_t$; \quad $P_{uni}^{(k)} \leftarrow s_{uni} \cdot p^{raw, (k)}_t$
    \State Pre-compute targets: $L_{ref} \leftarrow \mathcal{L}(P_{ind}^{(k)})$
\EndFor

\State \textbf{Phase 2: SQP Optimization}
\State Initialize $q_0 \leftarrow q_{nom}$
\For{$t = 1 \to T$}
    \State Detect interaction graph $\mathcal{E}_{inter}$ using $P_{uni}$
    \State Let $q_{prev} \leftarrow q_{t-1}$
    
    \State \textbf{Formulate QP Subproblem (Solve for $\Delta q$):}
    \State \textit{Objective:} $\min_{\Delta q} \mathcal{J}(q_{prev} + \Delta q)$
    \State \quad $= w_{self}(\|\mathcal{L}(q) \!-\! L_{ref}\|^2 \!+\! \lambda_{rot}\|\Delta \theta\|^2)$
    \State \quad $+ w_{inter} \sum_{(i,j)\in \mathcal{E}} \omega_{ij} \|\Delta p_{ij} - \Delta p^{uni}_{ij}\|^2$
    \State \quad $+ w_{reg} \|\Delta q\|^2$
    
    \State \textit{Subject to Constraints:}
    \State \quad 1) Limits: $q^{min} \le q_{prev} + \Delta q \le q^{max}$
    \State \quad 2) Collision: $J_{col} \Delta q \ge -\phi(q_{prev}) - \epsilon_{safe}$
    \State \quad 3) Contact: $\| J_{foot} \Delta q \| \le \epsilon_{stick}$ \textbf{if} foot\_contact
    \State \quad 4) Trust Region: $\| \Delta q \|_2 \le \delta$
    
    \State $\Delta q^* \leftarrow \text{OSQP}(\text{Objective}, \text{Constraints})$
    \State Update: $q_t \leftarrow q_{prev} + \Delta q^*$
    \State Append $q_t$ to $Q^{rob}$
\EndFor
\State \Return $Q^{rob}$
\end{algorithmic}
\end{algorithm}

\subsubsection{Topological Graph Visualization}
\label{subsubsec:iamr_topo}

\begin{figure}[t]
  \centering
  \includegraphics[width=0.5\linewidth]{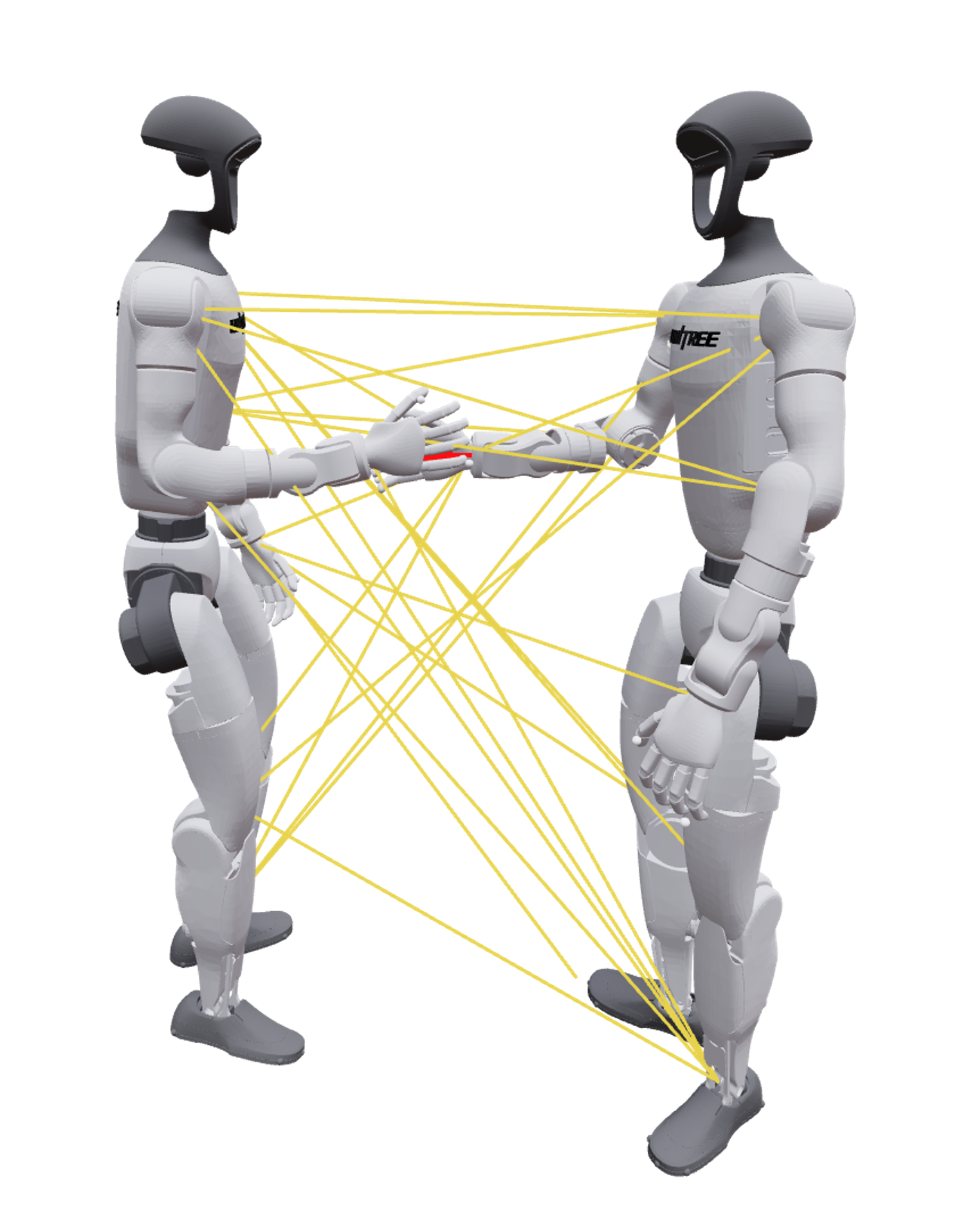}
  \caption{\textbf{Visualization of Topological Interaction Priors.} 
  We illustrate the extracted graph structures on a representative interaction task, where \textbf{yellow edges} denote spatial interaction constraints and \textbf{red edges} indicate physical contacts.}
  \label{fig:topo_vis}
\end{figure}

To provide an intuitive understanding of the topological priors, we visualize the extracted graph structures using a representative interaction case (e.g., a handshake task), as shown in Fig.~\ref{fig:topo_vis}.
The visualization highlights two distinct connectivity types used by IAMR:
\begin{itemize}
    \item \textbf{The Interaction Graph (Yellow Edges):} Bridges the keypoints of the two agents based on the Unified Manifold. It captures the \textit{spatial intent} and proximity required for coordination.
    \item \textbf{The Contact Graph (Red Edges):} Highlights active physical collision links between specific robot bodies. This explicitly encodes the \textit{physical coupling} state.
\end{itemize}
By explicitly modeling these connections, IAMR provides topological cues that guide the downstream policy to distinguish between required interaction and unwanted penetration.

\subsection{Details of IGRL}
\label{sec:append_igrl}

\subsubsection{Network Architecture}
\label{subsubsec:igrl_network}

Our policy $\pi_\theta(a_t|o_t)$ employs a hierarchical encoder-decoder architecture designed to process heterogeneous temporal data. The network input is composed of three semantic groups, which are processed by specialized encoders before being fused for action generation.

\vspace{5pt}
\noindent \textbf{Observation Space \& Inputs} \\
To capture complex coupled dynamics, the policy inputs are organized into two distinct temporal streams: a \textbf{Future Window} ($t+1 \dots t+20$) providing feedforward intent, and a \textbf{History Window} ($t-19 \dots t$) providing feedback control states.

\begin{itemize}
    \item \textbf{Future Reference ($o_{fut} \in \mathbb{R}^{93}$ per step):} Encodes the look-ahead trajectory for motion planning.
    \begin{itemize}
        \item \textit{Self Future ($\in \mathbb{R}^{64}$):} Contains the reference joint positions ($\in \mathbb{R}^{29}$) and velocities ($\in \mathbb{R}^{29}$), along with the reference root orientation ($\in \mathbb{R}^{6}$). The orientation is represented by the first two columns of the rotation matrix (Rot6D).
        \item \textit{Partner Future ($\in \mathbb{R}^{29}$):} Contains the partner's reference joint positions to anticipate collaborative intent.
    \end{itemize}

    \item \textbf{History Observation ($o_{hist} \in \mathbb{R}^{239}$ per step):} Aggregates the agent's proprioception and perception of the peer.
    \begin{itemize}
        \item \textbf{Proprioception ($o_{prop} \in \mathbb{R}^{157}$):}
        \begin{itemize}
            \item \textit{Tracking State ($\in \mathbb{R}^{64}$):} Includes the current step's reference joint positions and velocities, combined with the root orientation error ($\in \mathbb{R}^{6}$). The orientation error is computed from the first two columns of the rotation error matrix ($R_{des}^T R_{cur}$).
            \item \textit{Physical State ($\in \mathbb{R}^{93}$):} Includes projected gravity ($\in \mathbb{R}^{3}$), base angular velocity ($\in \mathbb{R}^{3}$), joint positions ($\in \mathbb{R}^{29}$), joint velocities ($\in \mathbb{R}^{29}$), and previous actions ($\in \mathbb{R}^{29}$).
        \end{itemize}
        
        \item \textbf{Peer Perception ($o_{peer} \in \mathbb{R}^{82}$):}
        \begin{itemize}
            \item \textit{Partner State ($\in \mathbb{R}^{64}$):} Includes the partner's current reference joint positions ($\in \mathbb{R}^{29}$), actual joint positions ($\in \mathbb{R}^{29}$), and the partner's root orientation error ($\in \mathbb{R}^{6}$).
            \item \textit{Interaction Topology ($\in \mathbb{R}^{18}$):} Explicitly encodes the formation geometry by including both the \textit{reference} and \textit{simulation} relative transforms. Each transform consists of the relative position ($p_{rel} \in \mathbb{R}^3$) and orientation ($R_{rel} \in \mathbb{R}^6$) expressed in the ego-centric frame.
        \end{itemize}
    \end{itemize}
\end{itemize}

\vspace{5pt}
\noindent \textbf{Temporal Encoders (1D-CNN)} \\
We employ two separate 1D-Convolutional Neural Networks to extract features from the observation history and future trajectories. Distinct from standard implementations, we employ an \textit{Input Projection} layer to compress high-dimensional inputs before temporal convolution.
\begin{itemize}
    \item \textbf{Architecture:} Both the History and Future encoders share a dual-layer CNN structure configured as follows:
    \begin{itemize}
        \item \textit{Input Projection:} Linear mapping to latent dim $C=60$.
        \item \textit{Layer 1:} Conv1d($60 \to 40$, $k=6, s=2$), ELU.
        \item \textit{Layer 2:} Conv1d($40 \to 20$, $k=4, s=2$), ELU.
    \end{itemize}
    \item \textbf{Output:} The temporal features are flattened and projected to compact embeddings ($e_{hist} \in \mathbb{R}^{67}, e_{fut} \in \mathbb{R}^{64}$). Uniquely, $e_{hist}$ incorporates an explicit estimation of the base linear velocity ($\hat{v}_{base} \in \mathbb{R}^3$) alongside the latent features. This design serves as a fusion of explicit physical estimation and implicit temporal representations, compensating for the lack of direct velocity observations in the real world.
\end{itemize}

\vspace{5pt}
\noindent \textbf{ Action Decoder (MLP)} \\
The encoded temporal features are concatenated with the current time-step observation and fed into a Multi-Layer Perceptron (MLP) to generate the action distribution.
\begin{itemize}
    \item \textbf{Structure:} Three hidden layers with $[512, 256, 128]$ units and ELU activation.
    \item \textbf{Output Head:} A linear layer outputs the mean $\mu_t \in \mathbb{R}^{29}$ of the Gaussian distribution for target joint positions. The standard deviation $\sigma$ is a learnable parameter initialized at 1.0.
\end{itemize}

\subsubsection{Reward Definitions}
\label{subsubsec:igrl_rewards}

The total reward $r_t$ is computed as a weighted sum of terms designed to balance kinematic fidelity with interaction plausibility, as detailed in Table~\ref{tab:rewards}. We prioritize interaction-centric objectives (e.g., relative geometry and contact) over individual tracking precision to encourage compliant multi-agent coupling.

\vspace{5pt}
\noindent \textbf{Interaction Graph Reward ($r_{inter}$)} \\
As formulated in the main text (Sec. III-B, \textit{Graph-based Rewards}), this term enforces geometric consistency by penalizing deviations in the interaction edges.
\begin{equation*}
    r_{inter} = \exp \left( - \frac{1}{\sigma_{inter}} \sum_{(i,j) \in \mathcal{E}} w_{ij} \cdot \| p_{ij}^{sim} - p_{ij}^{ref} \|^2 \right).
\end{equation*}
By inheriting the dynamic weights $w_{ij}$ from the IAMR module, the policy inherently learns to prioritize the same spatial topology (e.g., hand-shoulder proximity) as the optimized reference.

\vspace{5pt}
\noindent \textbf{Physical Contact Graph Reward ($r_{contact}$)} \\
Unlike kinematic tracking, physical interaction requires satisfying force constraints that are not present in motion datasets. We propose a graph-based formulation that handles contact nodes dynamically based on their reference status.
Leveraging the flexible node definition \cite{Wang_2025_CVPR}, we map simulation links to a set of abstract contact nodes $\mathcal{V}$ (e.g., palms, feet, pelvis). The reward is computed as a weighted sum of an \textit{active contact} term and an \textit{inactive} term:
\begin{equation*}
    r_{contact} = \lambda_{act} \cdot e^{-E_{act} / \sigma_c^2} + \lambda_{inact} \cdot e^{-E_{inact} / \sigma_c^2}.
\end{equation*}
The weights $\lambda_{act}$ and $\lambda_{inact}$ are adaptive, calculated as the ratio of active/inactive nodes in the current reference frame, ensuring balanced supervision across diverse contact phases.

\vspace{2pt}
\noindent \textit{i) Active Contact Error ($E_{act}$):}
For nodes where contact is expected ($k \in \mathcal{V}_{act}$), we define a hybrid error combining binary status consistency and continuous force regularization:
\begin{equation*}
    E_{act} = \sum_{k \in \mathcal{V}_{act}} \left( \beta \| C_{k}^{sim} - 1 \| + (1-\beta) \mathcal{L}_{force}(f_k^{sim}) \right).
\end{equation*}
where $C_k^{sim} \in \{0,1\}$ is the detected contact status. The force regularization term $\mathcal{L}_{force}$ penalizes forces outside the valid range $[F_{min}, F_{max}]$:
\begin{equation*}
    \mathcal{L}_{force}(f) = 
    \begin{cases} 
    1.0 - f/F_{min} & \text{if } f < F_{min} \text{ (Too Weak)} \\
    (f - F_{max})/F_{max} & \text{if } f > F_{max} \text{ (Too Strong)} \\
    0 & \text{otherwise (Valid)}
    \end{cases}.
\end{equation*}
This formulation explicitly guides the agent to exert sufficient force for stability ($>F_{min}$) while preventing explosive collisions ($<F_{max}$), solving the ambiguity of ``touching without force''.

\vspace{2pt}
\noindent \textit{ii) Inactive Contact Error ($E_{inact}$):}
For nodes where no contact is expected ($k \notin \mathcal{V}_{act}$), we strictly penalize any detected ``ghost interaction'':
\begin{equation*}
    E_{inact} = \sum_{k \notin \mathcal{V}_{act}} \| C_{k}^{sim} - 0 \|.
\end{equation*}

\begin{table*}[t]
\centering
\caption{Reward Terms and Weights used in IGRL}
\label{tab:rewards}
\small
\definecolor{Gray}{gray}{0.9}
\renewcommand{\arraystretch}{1.3}
\setlength{\tabcolsep}{3pt}

\begin{tabular*}{\textwidth}{@{\extracolsep{\fill}} l c p{6.8cm} p{7.5cm} }
\toprule
\textbf{Term} & \textbf{Weight} & \textbf{Equation} & \textbf{Description} \\ 
\midrule

\rowcolor{Gray} \multicolumn{4}{l}{\textbf{\textit{Interaction Graph Objectives}}} \\
Interact Edge & $1.5$ & $\exp \left( - \frac{1}{\sigma_{i}} \sum_{(i,j) \in \mathcal{E}} w_{ij} \| p_{ij}^{sim} - p_{ij}^{ref} \|^2 \right)$ & Enforces geometric consistency of interaction edges. \\

Contact & $1.0$ & $\lambda_{act} e^{-E_{act} / \sigma_c^2} + \lambda_{inact} e^{-E_{inact} / \sigma_c^2}$ \newline
{\footnotesize \textbf{where} $E_{act} = \sum_{k \in \mathcal{V}_{act}} (\beta \| C_{k}^{sim} - 1 \| + (1-\beta) \mathcal{L}_{f})$} \newline
{\footnotesize \textbf{and} \quad $E_{inact} = \sum_{k \notin \mathcal{V}_{act}} \| C_{k}^{sim} - 0 \|$} & Balances active contact enforcement (force constrained) and ghost interaction suppression. \\

\rowcolor{Gray} \multicolumn{4}{l}{\textbf{\textit{Motion Tracking Objectives}}} \\
Upper Pos & $1.0$ & $\exp(-\frac{1}{N_u} \sum_{k \in Upper} \|p_k^{sim} - p_k^{ref}\|^2 / \sigma^2_{pos})$ & Tracks Euclidean positions of upper body links. \\

Upper Ori & $1.0$ & $\exp(-\frac{1}{N_u} \sum_{k \in Upper} \|\log((R_k^{sim})^\top R_k^{ref})\|^2 / \sigma^2_{ori})$ & Tracks orientation of upper body links. \\

Upper Lin Vel & $1.0$ & $\exp(-\frac{1}{N_u} \sum_{k \in Upper} \|v_k^{sim} - v_k^{ref}\|^2 / \sigma^2_{vel})$ & Matches linear velocities of upper body. \\

Upper Ang Vel & $1.0$ & $\exp(-\frac{1}{N_u} \sum_{k \in Upper} \|\omega_k^{sim} - \omega_k^{ref}\|^2 / \sigma^2_{ang})$ & Matches angular velocities of upper body. \\

Lower Pos & $0.5$ & $\exp(-\frac{1}{N_l} \sum_{k \in Lower} \|p_k^{sim} - p_k^{ref}\|^2 / \sigma^2_{pos})$ & Tracks Euclidean positions of lower body links. \\

Lower Ori & $0.5$ & $\exp(-\frac{1}{N_l} \sum_{k \in Lower} \|\log((R_k^{sim})^\top R_k^{ref})\|^2 / \sigma^2_{ori})$ & Tracks orientation of lower body links. \\

Lower Lin Vel & $0.5$ & $\exp(-\frac{1}{N_l} \sum_{k \in Lower} \|v_k^{sim} - v_k^{ref}\|^2 / \sigma^2_{vel})$ & Matches linear velocities of lower body. \\

Lower Ang Vel & $0.5$ & $\exp(-\frac{1}{N_l} \sum_{k \in Lower} \|\omega_k^{sim} - \omega_k^{ref}\|^2 / \sigma^2_{ang})$ & Matches angular velocities of lower body. \\

Anchor Pos & $0.3$ & $\exp(-\|p_{root}^{sim} - p_{root}^{ref}\|^2 / \sigma^2_{root})$ & Tracks root position in world frame to prevent global drift. \\

Anchor Ori & $0.5$ & $\exp(-\|\log((R_{root}^{sim})^\top R_{root}^{ref})\|^2 / \sigma^2_{root\_ori})$ & Tracks root heading in world frame. \\

\rowcolor{Gray} \multicolumn{4}{l}{\textbf{\textit{Regularization \& Penalties}}} \\
Action Rate & $-0.3$ & $\|\mathbf{a}_t - \mathbf{a}_{t-1}\|^2$ & Penalizes action changes to ensure smooth control. \\

Feet Slip & $-0.5$ & $\sum_{k \in Feet} \mathbb{I}(contact_k) \cdot \|v_{k, xy}\|^2$ & Penalizes foot sliding velocity during ground contact. \\

Joint Limit & $-10.0$ & $\sum_{j} \max(0, |q_j| - q_{limit})$ & Penalizes violations of physical joint limits. \\

Torque & $10^{-4}$ & $-\|\tau\|^2$ & Prevents excessive torques. \\

\bottomrule
\end{tabular*}
\end{table*}

\subsubsection{Robust Training Strategy}
\label{subsubsec:igrl_robust}

To ensure transferability to the physical world and handle the complexity of coupled interaction phases, we implement a rigorous training protocol comprising error-aware curriculum-based adaptive sampling and extensive domain randomization.

\vspace{5pt}
\noindent \textbf{Curriculum-based Adaptive Sampling} \\
Standard Reference State Initialization (RSI) relies on sparse failure counts, which is insufficient for interaction tasks where ``survival'' does not imply ``success'' (e.g., maintaining balance but losing interaction geometry). We propose a continuous, error-aware sampling strategy defined as follows:
\begin{itemize}
    \item \textbf{Multi-Objective Error Landscape:} For each discretized motion bin $s$, we maintain a smoothed error vector $\mathbf{e}(s) = [e_{fail}, e_{track}, e_{inter}]^T$, incorporating failure signals, tracking errors, and interaction violations. To capture temporal causality—where failures stem from preceding suboptimal actions—we smooth these signals using a non-causal Gaussian kernel ($k=3$) implemented via 1D convolution.
    \item \textbf{Dynamic Probability:} The sampling probability $P(s)$ is a convex combination of uniform exploration ($\eta=0.05$) and error-weighted exploitation, modulated by curriculum weights $\boldsymbol{\alpha}$:
    \begin{equation*}
        P(s) = \eta \frac{1}{S} + (1-\eta) \sum_{k} \alpha_k(\bar{L}_{max}) \frac{e_k(s)}{\sum_{j} e_k(j)}.
    \end{equation*}
    \item \textbf{Curriculum Schedule:} The weight vector $\boldsymbol{\alpha}$ evolves based on the maximum moving average episode length $\bar{L}_{max}$ to ensure monotonic progress:
    \begin{itemize}
        \item \textit{Stability Phase ($\bar{L}_{max} < 350$):} Weights are fixed at $\boldsymbol{\alpha}_{init} = [0.8, 0.1, 0.1]$. The policy prioritizes states leading to terminal failures to learn basic balance capabilities.
        \item \textit{Transition Phase ($350 \le \bar{L}_{max} < 500$):} We perform \textbf{linear interpolation} between $\boldsymbol{\alpha}_{init}$ and the target weights $\boldsymbol{\alpha}_{target} = [0.05, 0.30, 0.65]$. This gradually shifts the focus from avoiding failures to minimizing tracking and interaction errors as the agent gains proficiency.
        \item \textit{Precision Phase ($\bar{L}_{max} \ge 500$):} Weights stabilize at $\boldsymbol{\alpha}_{target}$. The sampling strictly targets complex interaction phases that impose high control complexity despite low failure probability.
    \end{itemize}
\end{itemize}

\vspace{5pt}
\noindent \textbf{Dual-Agent Domain Randomization} \\
To bridge the Sim-to-Real gap, we introduce perturbations targeting both physical dynamics and the specific challenges of distributed multi-agent communication, as detailed in Table~\ref{tab:dr_params}.

\noindent \textbf{Communication \& Initialization Strategy.}
Beyond standard dynamics randomization, we implement specific strategies for dual-agent coordination:
\begin{itemize}
    \item \textbf{Communication Degradation:} We explicitly model the latency in distributed systems. By training with randomized delays ($20\sim60$ms) for both peer proprioception (wireless transmission) and relocalization (vision processing), the policy learns to be robust against asynchronous data reception.
    \item \textbf{Initial State Perturbation:} To handle calibration errors, we initialize episodes with random offsets in root position ($\pm 5$cm) and orientation ($\pm 0.2$rad). This forces the policy to learn active recovery behaviors from suboptimal relative configurations immediately upon activation.
\end{itemize}

\begin{table}[t]
\centering
\caption{Domain Randomization Parameters}
\label{tab:dr_params}
\small

\definecolor{Gray}{gray}{0.9}

\renewcommand{\arraystretch}{1.2}
\setlength{\tabcolsep}{0pt}

\begin{tabular*}{\columnwidth}{@{\extracolsep{\fill}} @{\hspace{4pt}} l p{5.2cm} @{\hspace{4pt}} }
\toprule
\textbf{Term} & \textbf{Value} \\ 
\midrule

\rowcolor{Gray} \multicolumn{2}{l}{\textbf{\textit{Dynamics Randomization}}} \\
\hspace{1em}Link Mass & $\mathcal{U}[0.9, 1.1] \times \text{default}$ (per link) \\
\hspace{1em}CoM Offset (Torso) & $\Delta x, y, z \in \mathcal{U}[-0.05, 0.05]$ m \\
\hspace{1em}Joint Friction & Static: $\mathcal{U}[0.3, 2.0]$, \newline Dynamic: $\mathcal{U}[0.3, 1.6]$ \\
\hspace{1em}Actuator Gains & Stiffness/Damping: $\mathcal{U}[0.9, 1.1]$ \\
\hspace{1em}Restitution & $\mathcal{U}[0.0, 0.8]$ (Ground contact) \\
\hspace{1em}Default Joint Pos & $\Delta \theta_0 \sim \mathcal{U}[-0.01, 0.01]$ rad (Calibration) \\
\hspace{1em}Control Delay & $\mathcal{U}[0, 15]$ ms \\

\rowcolor{Gray} \multicolumn{2}{l}{\textbf{\textit{External Perturbations}}} \\
\hspace{1em}Robot Push (Linear) & $v_{x,y} \sim \mathcal{U}[-0.4, 0.4]$ m/s, \newline $v_z \sim \mathcal{U}[-0.16, 0.16]$ m/s \\
\hspace{1em}Robot Push (Angular) & $\omega_{x,y} \sim \mathcal{U}[-0.4, 0.4]$ rad/s, \newline $\omega_z \sim \mathcal{U}[-0.64, 0.64]$ rad/s \\
\hspace{1em}Push Interval & Applied every $1.0 \sim 3.0$ s \\
\hspace{1em}Initial Pose Offset & $\Delta pos \in [-5, 5]$ cm, \newline $\Delta yaw \in [-0.2, 0.2]$ rad \\

\rowcolor{Gray} \multicolumn{2}{l}{\textbf{\textit{Communication Degradation}}} \\
\hspace{1em}Peer Latency & Proprioception \& Relocalization: \newline $\mathcal{U}[20, 60]$ ms \\

\bottomrule
\end{tabular*}
\vspace{-10pt}
\end{table}

\subsection{Experimental Setup \& Metrics}
\label{sec:append_exp}

% ...

\subsubsection{Baseline Implementation Details}
\label{subsubsec:exp_baselines}

To strictly validate our contributions, we benchmark our framework against two sets of baselines: kinematic retargeting methods (answering Q1) and dynamic policy learning variants (answering Q2).

\vspace{5pt}
\noindent \textbf{Retargeting Baselines (Kinematic Level)} \\
All baselines utilize the same source motion data and undergo identical skeletal scaling pre-processing to ensure fair comparison.
\begin{itemize}
    \item \textbf{GMR (General Motion Retargeting)~\cite{araujo2025retargeting}:} A standard Cartesian-space optimization method. It treats the two humanoids as isolated entities, optimizing joint angles to minimize the tracking error of individual keypoints' \textbf{positions and rotations} (e.g., end-effectors, pelvis) relative to the source motion. It does not include any interaction-aware constraints.
    
    \item \textbf{OR (OmniRetarget)~\cite{yang2025omniretarget}:} A retargeting method utilizing Interaction Mesh (I-Mesh) to preserve geometric topology, originally designed for Human-Object Interaction (HOI) scenarios. In our dual-agent setting, we apply OR to each agent independently. While it preserves individual body topology, it lacks mechanisms to model or preserve inter-agent spatial relationships.

    \item \textbf{DOR (Dual-OmniRetarget):} A strong baseline we constructed by extending OmniRetarget to the multi-agent setting. We construct a holistic interaction mesh that encompasses both robot bodies, creating ``cross-agent'' edges between proximal body parts. However, by treating the dual-agent system as a single unified mesh, it fails to distinguish between critical inter-agent interactions and individual self-motion, leading to an optimization that lacks focus on core interaction information.
\end{itemize}

\vspace{5pt}
\noindent \textbf{Policy Learning Baselines (Dynamic Level)} \\
To isolate the contribution of specific components in our IGRL framework, we compare against the following variants. All variants are trained using the same PPO hyperparameters and network architecture unless specified otherwise.

\begin{itemize}
    \item \textbf{Single Agent:} Represents the ``Status Quo'' approach. This policy treats the peer robot merely as a dynamic obstacle or ignores it entirely. \\
    \textit{Implementation:} The peer observation $o_{peer}$ is masked, and the reward function includes only standard tracking terms ($r_{track}$) without any interaction ($r_{inter}$) or contact ($r_{contact}$) objectives. The agent is rewarded solely for tracking its own retargeted reference.
    
    \item \textbf{Ours w/o Peer Obs:} Evaluates the necessity of explicit inter-agent perception. \\
    \textit{Implementation:} The policy architecture is identical to IGRL, but the peer observation stream $o_{peer}$ is zeroed out. The reward function remains full (including interaction rewards), forcing the agent to infer interaction requirements solely from its own proprioception and the reference motion.
    
    \item \textbf{Ours w/o Contact Rew:} Evaluates the contribution of the Physical Contact Graph. \\
    \textit{Implementation:} The policy is trained with the full observation space, but the contact reward $r_{contact}$ weight is set to zero. This tests whether kinematic tracking alone is sufficient to establish stable physical coupling.
    
    \item \textbf{Ours w/o Interaction Rew:} Evaluates the contribution of the Spatial Interaction Graph. \\
    \textit{Implementation:} The interaction graph reward $r_{inter}$ is removed. The agent relies purely on standard position/rotation tracking rewards to maintain formation. This tests the importance of explicitly optimizing relative topology for resolving geometric ambiguities.
\end{itemize}

\subsubsection{Evaluation Metric Implementation Details}
\label{subsubsec:exp_metrics}

We employ specific metrics for each component of our framework to evaluate kinematic quality and dynamic performance respectively.

\vspace{5pt}
\noindent \textbf{Retargeting Evaluation Metrics (Q1)} \\
We evaluate retargeting quality from three complementary aspects: physical feasibility, interaction fidelity, and downstream utility.

\begin{itemize}
    \item \textbf{Inter-Penetration Rate (IPR) \& Max Penetration Depth (MPD):} 
    Measure physical feasibility by reporting the percentage of frames exhibiting inter-agent penetration and the maximum penetration depth across the sequence. Lower values indicate safer motion.

    \item \textbf{Interaction Edge Error (IEE):} 
    Quantifies geometric interaction fidelity as the normalized L2 distance between retargeted interaction edges and the scaled ground-truth edges.

    \item \textbf{Contact F1 Score:} 
    Evaluates binary contact accuracy against ground-truth interaction edges following \cite{liu2025takes,huang2026learning}. We report F1 scores under \textbf{Strict} ($\tau < 0.2$\,m) and \textbf{Loose} ($\tau < 0.4$\,m) contact settings to assess precision at different scales.

    \item \textbf{Downstream Success Rate (DSR):} 
    Serves as a proxy for physical learnability. It is defined as the percentage of RL rollouts where the agent can track the retargeted reference while maintaining the interaction structure (i.e., maintaining an IEE deviation $< 20\%$ relative to the scaled ground truth).
\end{itemize}

\vspace{5pt}
\noindent \textbf{Policy Learning Evaluation Metrics (Q2)} \\
To assess the robustness and fidelity of the learned control policy in dynamic environments, we evaluate interaction and contact performance with respect to the retargeted reference trajectories.

\begin{itemize}
    \item \textbf{Interaction Performance (ISR \& IEE):} 
    Measures interaction fidelity by computing the distance-weighted relative error between simulated interaction edges and the retargeted reference edges. We report the mean \textbf{Interaction Edge Error (IEE)} and the \textbf{Interaction Success Rate (ISR)}, defined as the percentage of steps where the IEE is kept within a strict tolerance ($< 10\%$).

    \item \textbf{Contact Performance (CSR \& CER):} 
    Evaluates physical contact fidelity relative to the retargeted reference contacts. The \textbf{Contact Error Rate (CER)} measures the rate of violations of required contact constraints, while the \textbf{Contact Success Rate (CSR)} reports the percentage of steps where sufficient reference contacts ($>80\%$) are correctly recalled by the policy.
\end{itemize}

\subsubsection{Sim-to-Real Hardware}
\label{subsubsec:exp_hardware}

We validate our approach on the Unitree G1 humanoid robot platform. To bridge the gap between ideal simulation states and real-world noisy sensor data, we implement a fully onboard perception and control stack written in C++ for real-time performance.

\vspace{5pt}
\noindent \textbf{Robot Platform \& Compute.}
The Unitree G1 (approx. 1.3\,m height, 29 DoF) serves as our experimental testbed. Unlike simulation where state information is privileged, all computations—including state estimation, policy inference, and low-level control—are performed onboard the robot's internal CPU. We utilize ONNX Runtime to execute the trained policies, achieving an inference latency of less than 3\,ms, ensuring a stable 50\,Hz control loop.

\vspace{5pt}
\noindent \textbf{Hierarchical Localization System.}
Precise relative localization is a prerequisite for our \textit{Interaction Graph} reward calculation. We develop a robust, coarse-to-fine localization framework fusing LiDAR and IMU data:
\begin{itemize}
    \item \textbf{High-Frequency Odometry:} We utilize \textbf{Point-LIO} \cite{he2023point}, a robust LiDAR-Inertial Odometry framework, to provide high-bandwidth ($10$\,Hz) state estimation robust to aggressive motions and vibrations.
    \item \textbf{Global Initialization (Coarse):} To handle the ``cold start'' problem and global drift, we implement a neural registration service based on \textbf{GeoTransformer}. This module extracts superpoint features to perform global registration between the current scan and a pre-built point cloud map, providing a reliable initial pose guess.
    \item \textbf{Real-Time Tracking (Fine):} During operation, a C++ relocalization node refines the pose using \textbf{GICP} (Generalized Iterative Closest Point) \cite{segal2009generalized}. A Kalman Filter fuses the high-frequency relative odometry from Point-LIO with the low-frequency global pose corrections from GICP to output smooth, drift-free global states.
\end{itemize}

\vspace{5pt}
\noindent \textbf{Control Architecture.}
The deployment system operates on a dual-frequency hierarchy to match the simulation setup:
\begin{itemize}
    \item \textbf{High-Level Policy (50\,Hz):} The motion tracking policy constructs observations based on the fused state estimates and computes desired joint positions.
    \item \textbf{Low-Level Controller (500\,Hz):} A real-time PD controller converts these targets into motor torques, enforcing the physical compliance required for safe interaction.
\end{itemize}

\vspace{5pt}
\noindent \textbf{Multi-Agent Synchronization.}
To enable coordinated interaction, the agents exchange their estimated global poses and full peer observations ($o_{peer}$) via the \textbf{LCM} (Lightweight Communications and Marshalling) protocol~\cite{huang1998lcm}.

\end{document}